\documentclass[runningheads]{llncs}
\usepackage{graphicx}
\usepackage{comment}
\usepackage{amsmath,amssymb} 
\usepackage{color}

\begin{document}
\pagestyle{headings}
\mainmatter
\def\ECCVSubNumber{3649}  

\title{DELTAS: Depth Estimation by Learning Triangulation And densification of Sparse points} 


\titlerunning{Depth by Triangulation and Densification}
%
\author{Ayan Sinha\inst{1} \and
Zak Murez\inst{1} \and
James Bartolozzi \inst{*} \and Vijay Badrinarayanan \inst{2*} \and Andrew Rabinovich \inst{3*}}
\authorrunning{A. Sinha et al.}
%
\institute{Magic Leap Inc., CA, USA \email{\{asinha,zmurez\}@magicleap.com}\\
*Work done at Magic Leap \email{bartolozzij@gmail.com} \and
Wayve.ai, London, UK \email{vijay@wayve.ai} 
\and
InsideIQ Inc., CA, USA \email{andrew@insideiq.team}}
\maketitle

\begin{abstract}
Multi-view stereo (MVS) is the golden mean between the accuracy of active depth sensing and the practicality of monocular depth estimation. Cost volume based approaches employing 3D convolutional neural networks (CNNs) have considerably improved the accuracy of MVS systems. However, this accuracy comes at a high computational cost which impedes practical adoption. Distinct from cost volume approaches, we propose an efficient depth estimation approach by first (a) detecting and evaluating descriptors for interest points, then (b) learning to match and triangulate a small set of interest points, and finally (c) densifying this sparse set of 3D points using CNNs. An end-to-end network efficiently performs all three steps within a deep learning framework and trained with intermediate 2D image and 3D geometric supervision, along with depth supervision.  Crucially, our first step complements pose estimation using interest point detection and descriptor learning. We demonstrate state-of-the-art results on depth estimation with lower compute for different scene lengths. Furthermore, our method generalizes to newer environments and the descriptors output by our network compare favorably to strong baselines. Code is available at \url{https://github.com/magicleap/DELTAS}  










\end{abstract}

\section{Motivation}

Depth sensing is crucial for a wide range of applications ranging from Augmented Reality (AR)/ Virtual Reality (VR) to autonomous driving. Estimating depth can be broadly divided into classes: active and passive sensing. Active sensing techniques include LiDAR, structured-light and time-of-flight (ToF) cameras, whereas depth estimation using a monocular camera or stereopsis of an array of cameras is termed passive sensing. Active sensors are currently the de-facto standard of applications requiring depth sensing due to good accuracy and low latency in varied environments \cite{zhang2012microsoft}. However, active sensors have their own of limitation. LiDARs are prohibitively expensive and provide sparse measurements. Structured-light and ToF depth cameras have limited range and completeness due to the physics of light transport. Furthermore, they are power hungry and inhibit mobility critical for AR/VR applications on wearables. Consequently, computer vision researchers have pursued passive sensing techniques  as a ubiquitous, cost-effective and energy-efficient alternative to active sensors \cite{Ma2017SparseToDense}. 

Passive depth sensing using a stereo cameras requires a large baseline and careful calibration for accurate depth estimation \cite{chang2018pyramid}. A large baseline is infeasible for mobile devices like phones and wearables. An alternative is to use MVS techniques for a moving monocular camera to estimate depth. MVS generally refers to the problem of reconstructing 3D scene structure from multiple images with known camera poses and intrinsics \cite{hartley2003multiple}. The unconstrained nature of camera motion alleviates the baseline limitation of stereo-rigs, and the algorithm benefits from multiple observations of the same scene from continuously varying viewpoints \cite{hou2019multi}. However, camera motion also makes depth estimation more challenging relative to rigid stereo-rigs due to pose uncertainty and added complexity of motion artifacts. Most MVS approaches involve building a 3D cost volume, usually with a plane sweep stereo approach \cite{yao2018mvsnet,huang2018deepmvs}. Accurate depth estimation using MVS rely on 3D convolutions on the cost volume, which is both memory as well as computationally expensive, scaling cubically with the resolution. Furthermore, redundant compute is added by ignoring useful image-level properties such as interest points and their descriptors, which are a necessary precursor to camera pose estimation, and hence, any MVS technique.  This increases the overall cost and energy requirements for passive sensing.  

Passive sensing using a single image is fundamentally unreliable due to scale ambiguity in 2D images. Deep learning based monocular depth estimation approaches formulate the problem as depth regression \cite{eigen2015predicting,fu2018deep} and have reduced the performance gap to those of active sensors \cite{lee2019big,lasinger2019towards}, but still far from being practical. Recently, sparse-to-dense depth estimation approaches have been proposed to remove the scale ambiguity and improve robustness of monocular depth estimation \cite{Ma2017SparseToDense}. Indeed, recent sparse-to-dense approaches with less than $0.5\%$ depth samples have accuracy comparable to active sensors, with higher range and completeness \cite{chen2018estimating} . However, these approaches assume accurate or seed depth samples from an active sensor which is limiting. The alternative is to use the sparse 3D landmarks output from the best performing algorithms for Simultaneous Localization and Mapping (SLAM) \cite{mur2015orb} or Visual Inertial Odometry (VIO) \cite{nister2004visual}. However, using depth evaluated from these sparse landmarks in lieu of depth from active sensors, significantly degrades performance \cite{deep2018funk}. This is not surprising as the learnt sparse-to-dense network ignores potentially useful cues, structured noise and biases present in SLAM or VIO algorithm. 

Here we propose to learn the sparse 3D landmarks in conjunction with the sparse to dense formulation in an end-to-end manner so as to (a) remove dependence on a cost volume in the MVS technique,thus, significantly reducing compute, (b) complement camera pose estimation using sparse VIO or SLAM by reusing detected interest points and descriptors, (c) utilize  geometry-based MVS concepts to guide the algorithm and improve the interpretability,  and (d) benefit from the accuracy and efficiency of sparse-to-dense techniques. Our network is a multitask model \cite{kendall2018multi}, comprised of an encoder-decoder structure composed on two encoders, one for RGB image and one for sparse depth image, and three decoders: one for interest point detection, one for descriptors and one for the dense depth prediction. We also contribute a differentiable module that efficiently triangulates points using geometric priors and forms the critical link between the interest point decoder, descriptor decoder, and the sparse depth encoder enabling end-to-end training.   
 
The rest of the paper is organized as follows. Section 2 discussed related work and Section 3 describes our approach. We perform experimental evaluation in Section 4, and finally conclusions and future work are presented in Section 5. 

\section{Related Work}

\subsubsection{Interest point detection and description:} Sparse feature based methods are standard for SLAM or VIO techniques due to their high speed and accuracy. The detect-then-describe approach is the most common approach to sparse feature extraction, wherein, interest points are detected and then described for a patch around the point. The descriptor encapsulates higher level information, which are missed by typical low-level interest points such as corners, blobs, etc. Prior to the deep learning revolution, classical systems like SIFT \cite{lowe2004distinctive} and ORB \cite{rublee2011orb} were ubiquitously used as descriptors for feature matching for low level vision tasks. Deep neural networks directly optimizing for the objective at hand have now replaced these hand engineered features across a wide array of applications. However, such an end-to-end network has remained elusive for SLAM \cite{murthy2019gradslam} due to the components being non-differentiable. General purpose descriptors learnt by methods such as SuperPoint \cite{superpoint}, LIFT \cite{yi2016lift}, GIFT \cite{liu2019gift} aim to bridge the gap towards differentiable SLAM. 




\subsubsection{MVS:} MVS approaches either directly reconstruct a 3D volume or output a depth map which can be flexibly used for 3D reconstruction or other applications. Methods reconstructing 3D volumes \cite{yao2018mvsnet,chen2019point} are restricted to small spaces or isolated objects either due to the high memory load of operating in a 3D voxelized space \cite{riegler2017octnet,sinha2017surfnet}, or due to the difficulty of learning point representations in complex environments \cite{qi2017pointnet}. Here, we use multi-view images captured in indoor environments for depth estimation due to the versatility of depth map representation. This area has lately seen a lot of progress starting with DeepMVS \cite{huang2018deepmvs} which proposed a learnt patch matching approach. MVDepthNet \cite{wang2018mvdepthnet}, and DPSNet \cite{im2019dpsnet} build a cost volume for depth estimation. GP-MVSNet \cite{hou2019multi} built upon MVDepthNet to coherently fuse temporal information using gaussian processes. All these methods utilize the plane sweep algorithm during some stage of depth estimation, resulting in an accuracy vs efficiency trade-off. 

\subsubsection{Sparse to Dense Depth prediction:} Sparse-to-dense depth estimation has recently emerged as a way to supplement active depth sensors due to their range limitations when operating on a power budget, and to fill in depth in hard to detect regions such as dark or reflective objects. The first such approach was proposed by Ma et.al\cite{Ma2017SparseToDense}, and following work by Chen et. al. \cite{chen2018estimating} and \cite{deep2018funk} introduced innovations in the representation and network architecture. A convolutional spatial propagation module is  proposed in \cite{cheng2018depth} to in-fill the missing depth values. Self-supervised approaches \cite{godard2017unsupervised,garg2016unsupervised} have concurrently been explored for the sparse-to-dense problem \cite{ma2019self}. Recently, a learnable triangulation technique was proposed to learn human pose key-points \cite{iskakov2019learnable}. We leverage their algebraic triangulation module for the purpose of sparse reconstruction of 3D points.

\section{Method}

\begin{figure}[t]
\centering
\includegraphics[width=1.0\textwidth]{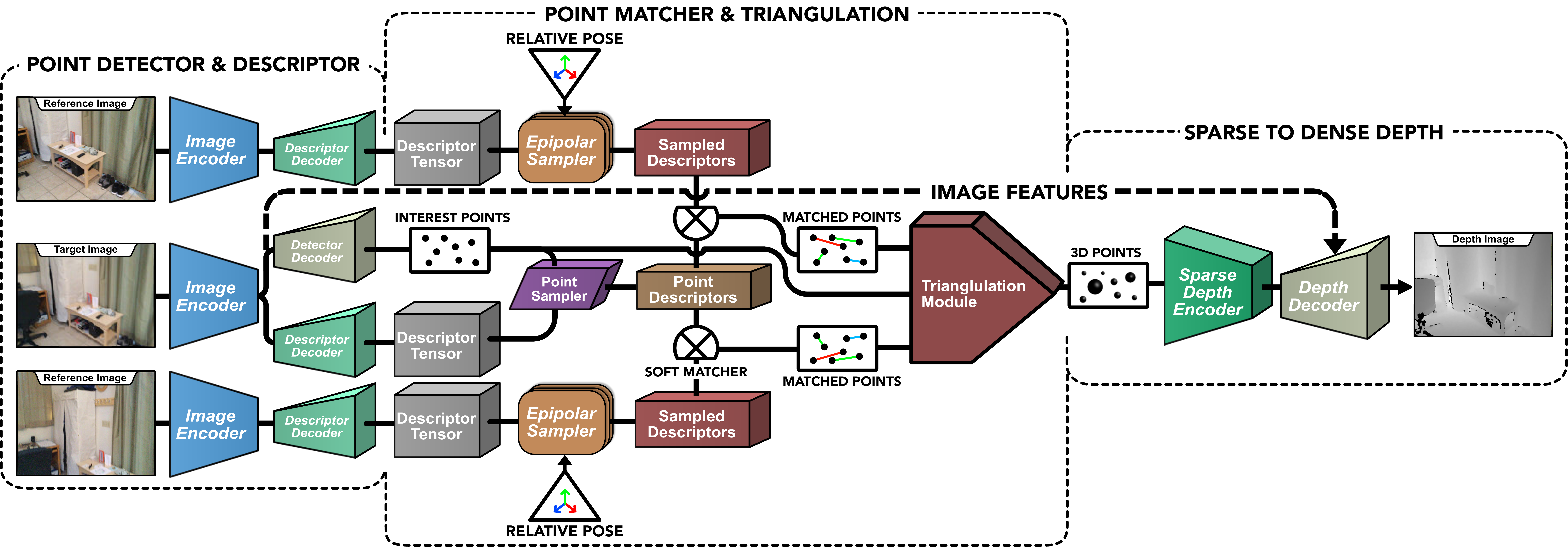}
\caption{End-to-end network for detection and description of interest points, matching and triangulation of the points and densification of 3D points for depth estimation.} 
\label{fig1}
\end{figure}

Our method can be broadly sub-divided into three steps as illustrated in Figure \ref{fig1} for a prototypical target image and two view-points. In the first step, the target or anchor image and the multi-view images are passed through a shared RGB encoder and descriptor decoder to output a descriptor field for each image. Interest points are also detected for the target or the anchor image. In the second step, the interest points in the anchor image in conjunction with the relative poses are used to determine the search space in the reference or auxiliary images from alternate view-points. Descriptors are sampled in the search space and are matched with descriptors for the interest points. Then, the matched key-points are triangulated using SVD and the output 3D points are used to create a sparse depth image. In the third and final step, the output feature maps for the sparse depth encoder and intermediate feature maps from the RGB encoder are collectively used to inform the depth decoder and output a dense depth image. Each of the three steps are described in greater detail below.  

\subsection{Interest point detector and descriptor}
\begin{figure}[t]
\centering
\includegraphics[width=1.0\textwidth]{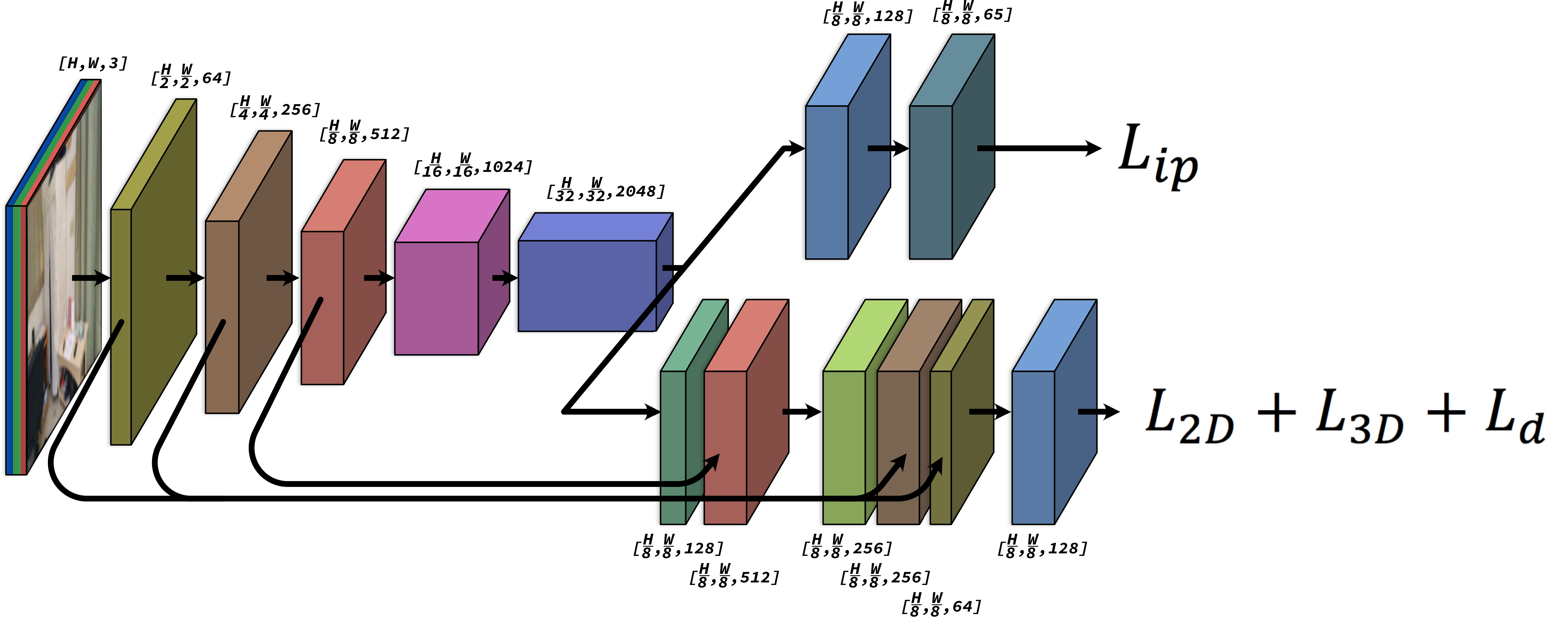}
\caption{ SuperPoint-like network with detector and descriptor heads. } 
\label{figsuperpoint}
\end{figure}

We adopt SuperPoint-like \cite{superpoint} formulation of a fully-convolutional neural network architecture which operates on a full-resolution image and produces interest point detection accompanied by fixed length descriptors. The model has a single, shared encoder to process and reduce the input image dimensionality. The feature maps from the encoder feed into two task- specific decoder “heads”, which learn weights for interest point detection and interest point description. This joint formulation of interest point detection and description in SuperPoint enables sharing compute for the detection and description tasks, as well as the down stream task of depth estimation. However, SuperPoint was trained on gray-scale images with focus on interest point detection and description for continuous pose estimation on high frame rate video streams, and hence, has a relatively shallow encoder. On the contrary, we are interested in image sequences with sufficient baseline, and consequently longer intervals between subsequent frames. Furthermore, SuperPoint's shallow backbone suitable for sparse point analysis has limited capacity for our downstream task of dense depth estimation. Hence, we replace the shallow backbone with a ResNet-50 \cite{he2016deep} encoder which balances efficiency and performance. The output resolution of the interest point detector decoder is identical to that of SuperPoint. In order to fuse fine and coarse level image information critical for point matching, we use a U-Net \cite{ronneberger2015u} like architecture for the descriptor decoder. This decoder outputs an N-dimensional descriptor tensor at $1/8^{th}$ the image resolution, similar to SuperPoint. The architecture is illustrated in Figure \ref{figsuperpoint}. We train the interest point detector network by distilling the output of the original SuperPoint network and the descriptors are trained by the matching formulation described below.   

\subsection{Point matching and triangulation}
\begin{figure}[t]
\centering
\includegraphics[width=1.0\textwidth]{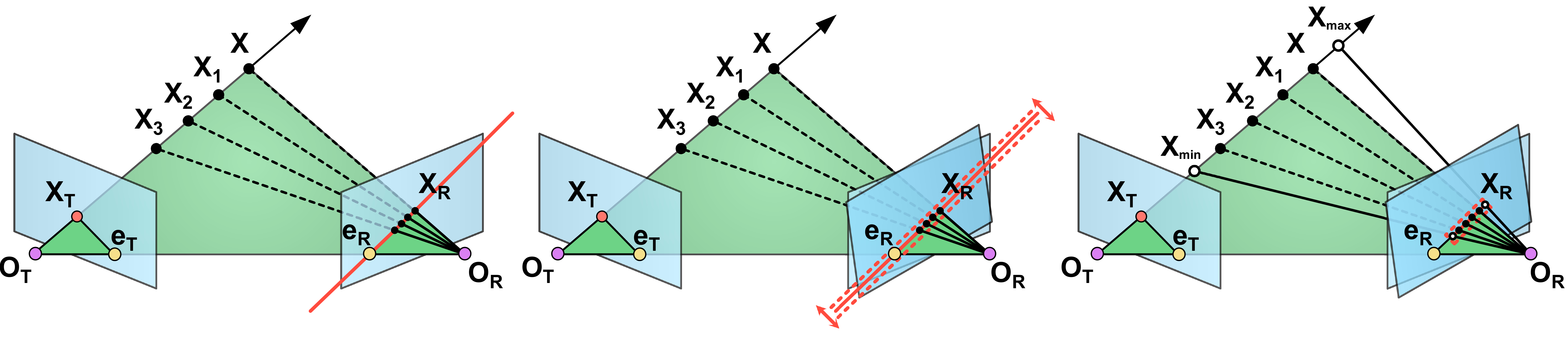}
\caption{Left: Epipolar sampling; Middle: Offset sampling due to relative pose error; Right: Constrained depth range sampling} 
\label{fig3}
\end{figure}

The previous step provides interest points for the anchor image and descriptors for all images, i.e., the anchor image and full set of auxiliary images. A naive approach will be to match descriptors of the interest points sampled from the descriptor field of the anchor image to all possible positions in each auxiliary image. However, this is computationally prohibitive. Hence, we invoke geometrical constraints to restrict the search space and improve efficiency. Using concepts from multi-view geometry, we only search along the epipolar line in the auxiliary images \cite{hartley2003multiple}. The epipolar line is determined using the fundamental matrix, $F$, using the relation $xFx^{T}=0$, where $x$ is the set of points in the image. The matched point is guaranteed to lie on the epipolar line in an ideal scenario as illustrated in Figure \ref{fig3} (Left). However, practical limitations to obtain perfect pose lead us to search along the epipolar line with a small fixed offset on either side; Figure \ref{fig3} (Middle). Furthermore, the epipolar line stretches for depth values from  $-\infty$  to $\infty$. We clamp the epipolar line to lie within feasible depth sensing range, and vary the sampling rate within this restricted range in order to obtain descriptor fields with the same output shape for implementation purposes, shown in Figure \ref{fig3} (Right). We use bilinear sampling to obtain the descriptors at the desired points in the descriptor field. The descriptor of each interest point is convolved with the descriptor field along its corresponding epipolar line for each image view-point:
\begin{equation}\label{advce}
  C_{j,k} = \hat D_{j} \ast D^k_{j},  \forall x \in \mathcal{E},
\end{equation}
where $\hat D$ is the descriptor field of the anchor image, $D^k$ is the descriptor field of the $k^{th}$ auxiliary image, and convolved over all sampled points $x$ along the clamped epipolar line $\mathcal{E}$ for point $j$.  This effectively provides a cross-correlation map \cite{bertinetto2016fully} between the descriptor field and interest point descriptors. High values in this map indicate potential key-point matches in the auxiliary images to the interest points in the anchor image. In practice, we add batch normalization \cite{ioffe2015batch} and ReLU non-linearity \cite{krizhevsky2012imagenet} to output $C_{j,k}$ in order to ease training. 

To obtain the 3D points, we follow the algebraic triangulation approach proposed in \cite{iskakov2019learnable}. We process each interest point $j$ independently of each other. The approach is built upon triangulating the 2D interest points along with the 2D positions obtained from the peak value in each cross correlation map. To estimate the 2D positions we first compute the softmax across the spatial axes:

\begin{equation}\label{advce}
  C^{'}_{j,k} =\exp(C_{j,k})/(\sum^W_{r_x=1} \sum_{r_y=1}^H\exp(C_{j,k}(r_x,r_y)),
\end{equation}
where, $C_{j,k}$ indicates the cross-correlation map for the $j^{th}$ inter-point and $k^{th}$ view, and $W,H$ are spatial dimensions of the epipolar search line. Then we calculate the 2D positions of the points as the center of mass of the corresponding cross-correlation maps, also termed soft-argmax operation:
\begin{equation}\label{advce1}
  x_{j,k} =\sum^W_{r_x=1} \sum_{r_y=1}^H r(x,y) (C^{'}_{j,k}(r(x,y))).
\end{equation}
The soft-argmax operation enables differentiable routing between the 2D position of the matched points $x_{j,k}$ and the cross-correlation maps $C_{j,k}$. We use the linear algebraic triangulation approach proposed in \cite{iskakov2019learnable} to estimate the 3D points from the matched 2D points $x_{j,k}$. Their method reduces the finding of the 3D coordinates of a point $z_j$ to solving the over-determined system of equations on homogeneous 3D coordinate vector of the point $\bar z$:


\begin{equation}\label{advce2}
  A_j \bar{z_j} =0, 
\end{equation}
where $A_j \in \mathcal{R}^{2k,4}$ is a matrix composed of the components from the full projection matrices and $x_{j,k}$. Different view-points may contribute unequally to the triangulation of a point due to occlusions and motion artifacts. Weighing the contributions equally leads to sub-optimal performance. The problem is solved in a differentiable way by adding weights $w_k$ to the coefficients of the matrix corresponding to different views:
\begin{equation}\label{advce2}
  (w_j A_j) \bar{z_j}=0. 
\end{equation}
The weights $w$ are set to be the max value in each cross-correlation map. This allows the contribution of the each camera view to be controlled by the quality of match, and low-confidence matches to be weighted less while triangulating the interest point. Note the confidence value of the interest points are set to be 1. The above equation is solved via differentiable Singular Value Decomposition (SVD) of the matrix $B = UDV^T$, from which $\bar{z}$ is set as the last column of $V$. The final non-homogeneous value of $z$ is obtained by dividing the homogeneous 3D coordinate vector $\bar{z}$ by its fourth coordinate: $z = \bar{z}/(\bar{z})_4$ \cite{iskakov2019learnable}.

\begin{figure}
\centering
\includegraphics[width=1.0\textwidth]{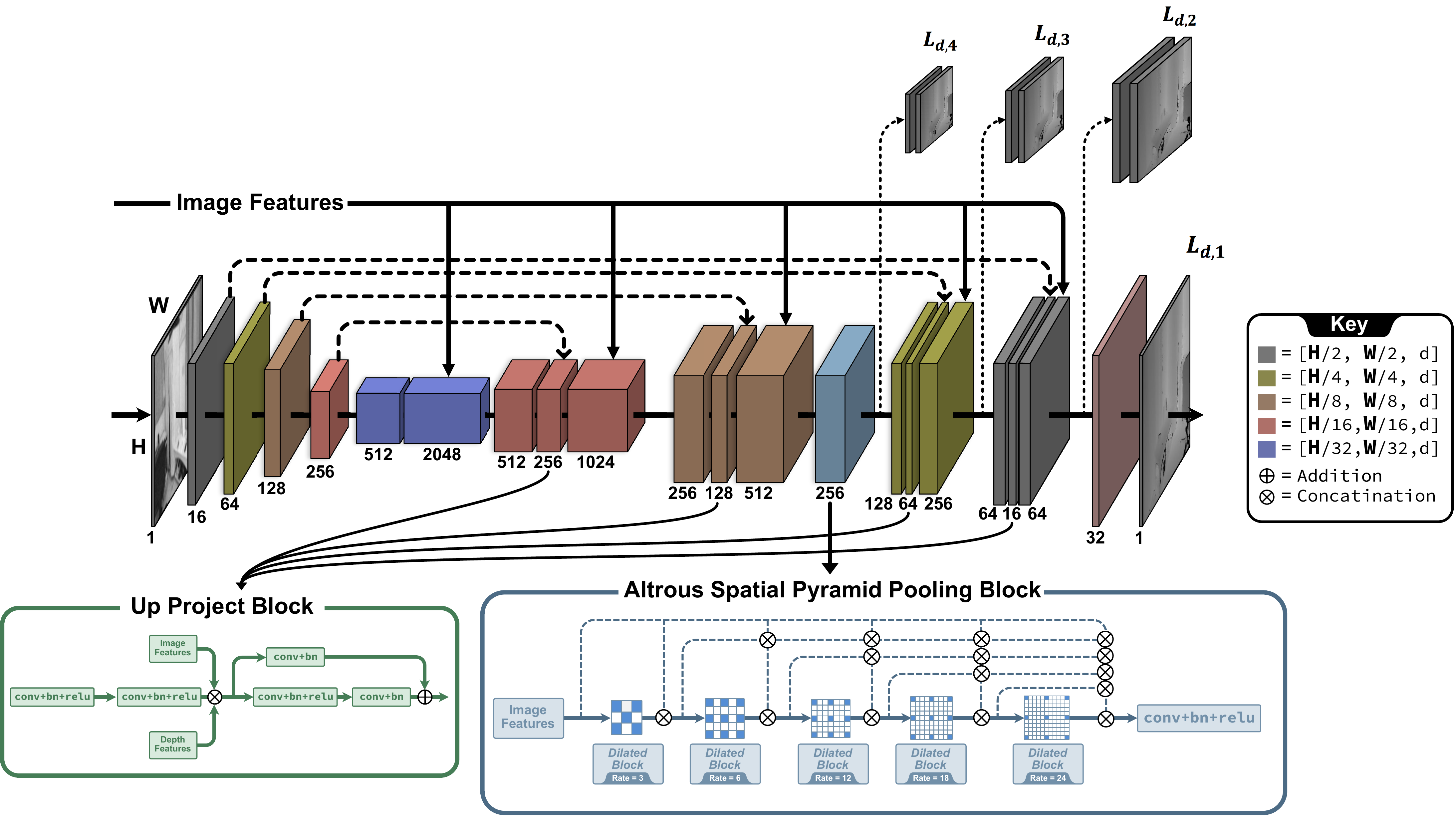}
\caption{Proposed sparse-to-dense network architecture showing the concatenation of image and sparse depth features. We use deep supervision over 4 image scales. The blocks below illustrate the upsampling and the altrous spatial pyramid pooling (ASPP) block.} 
\label{sparse2dense}
\end{figure}

\subsection{Densification of sparse depth points}
The interest-point detector network provides the 2D position of the points. The $z$ coordinate of the triangulated points provides the depth. We impute a sparse depth image of the same resolution as the input image with  depth of these sparse points. Note that the gradients can propagate from the sparse depth image back to the 3D key-points all the way to the input image. This is akin to switch unpooling in SegNet \cite{badrinarayanan2015segnet}. We pass the sparse depth image through an encoder network which is a narrower version of the image encoder network. Specifically, we use a ResNet-50 encoder with the channel widths after each layer to be $1/4^{th}$ of the image encoder. We concatenate these features with the features obtained from the image encoder. We use a U-net style decoder with intermediate feature maps from both the image as well as sparse depth encoder concatenated with the intermediate feature maps of the same resolution in the decoder, similar to  \cite{chen2018estimating}. We provide deep supervision over 4 scales \cite{lee2015deeply}. We also include a spatial pyramid pooling block to encourage feature mixing at different receptive field sizes \cite{he2015spatial,chen2017rethinking}. The details of the architecture are shown in the Figure \ref{sparse2dense}. 

\subsection{Overall training objective}
The entire network is trained with a combination of (a) cross entropy loss between the output tensor of the interest point detector decoder and ground truth interest point locations obtained from SuperPoint, (b) a smooth-L1 loss between the 2D points output after soft argmax and ground truth 2D point matches, (c) a smooth-L1 loss between the 3D points output after SVD triangulation and ground truth 3D points, (d) an edge aware smoothness loss on the output dense depth map, and (e) a smooth-L1 loss over multiple scales between the predicted dense depth map output and ground truth 3D depth map. The overall training objective is: 

\begin{equation}\label{advce2}
  L = w_{ip} L_{ip} + w_{2d} L_{2d} + w_{3d} L_{3d} + w_{sm} L_{sm} + \sum_i w_{d,i} L_{d,i}, 
\end{equation}
where $L_{ip}$ is the interest point detection loss, $L_{2d}$ is the 2D matching loss, $L_{3d}$ is the 3D triangulation loss, $L_{sm}$ is the smoothness loss, and $L_{d,i}$ is the depth estimation loss at scale $i$ for 4 different scales ranging from original image resolution to $1/16^{th}$ the image resolution. 

\section{Experimental Results}

\subsection{Implementation Details}

\subsubsection{Training:} Most MVS approaches are trained on the DEMON dataset \cite{ummenhofer2017demon}. However, the DEMON dataset mostly contains pairs of images with the associated depth and pose information. Relative confidence estimation is crucial to accurate triangulation in our algorithm, and needs sequences of length three or greater in order to estimate the confidence accurately and holistically triangulate an interest point. Hence, we diverge from traditional datasets for MVS depth estimation, and instead use ScanNet \cite{dai2017scannet}. ScanNet is an RGB-D video dataset containing 2.5 million views in more than 1500 scans, annotated with 3D camera poses, surface reconstructions, and instance-level semantic segmentations. Three views from a scan at a fixed interval of 20 frames along with the pose and depth information forms a training data point in our method. The target frame is passed through SuperPoint in order to detect interest points, which are then distilled using the loss $L_{ip}$ while training our network. We use the depth images to determine ground truth 2D matches, and unproject the depth to determine the ground truth 3D points. We train our model for ~100K iterations using PyTorch framework with batch-size of 24 and ADAM optimizer with learning rate 0.0001 ($\beta_1$ = 0.9, $\beta_2$ = 0.999), which takes about 3 days across 4 Nvidia Titan RTX GPUs. . We fix the resolution of the image to be qVGA (240$\times$320) and number of interest points to be 512 in each image with at most half the interest points chosen from the interest point detector thresholded at 0.0005, and the rest of the points chosen randomly from the image. Choosing random points ensures uniform distribution of sparse points in the image and helps the densification process. We set the length of the sampled descriptors along the epipolar line to be 100, albeit, we found that the matching is robust even for lengths as small as 25. We set the range of depth estimation to be between 0.5 and 10 meters, as common for indoor environments. We empirically set the weights to be [0.1,1.0,2.0,1.0,2.0] for $w_{ip}, w_{2d}, w_{3d}, w_{sm}, w_{d,1}$, respectively. We damp $w_{d,1}$ by a factor of 0.7 for each subsequent scale. 

\subsubsection{Evaluation:} The ScanNet test set consists of 100 scans of unique scenes different for the 707 scenes in the training dataset. We first evaluate the performance of our detector and descriptor decoder for the purpose of pose estimation on ScanNet. We use the evaluation protocol and metrics proposed in SuperPoint, namely the mean localization error (MLE), the matching score (MScore), repeatability (Rep) and the fraction of correct pose estimated using descriptor matches and PnP algorithm at $5^{\circ}$ threshold for rotation and and 5 cm for translation. We compare against SuperPoint, SIFT, ORB and SURF at a NMS threshold of 3 pixels for Rep, MLE, and MScore as suggested in the SuperPoint paper.
Next, we use standard metrics to quantitatively measure the quality of our estimated depth: : absolute relative error (Abs Rel), absolute difference error (Abs diff), square relative error (Sq Rel), root mean square error and its log scale (RMSE and RMSE log) and inlier ratios ($\delta < 1.25^i$ where $i \in {1, 2, 3}$). Note higher values for inlier ratios are desirable, whereas all other metrics warrant lower values. 

We compare our method to recent deep learning approaches for MVS: (a) DPSNet: Deep plane sweep approach, (b) MVDepthNet: Multi-view depth net, and (c) GPMVSNet temporal non-parametric fusion approach using Gaussian processes. Note that these methods perform much better than traditional geometry-based stereo algorithms. Our primary results are on sequences of length 3, but we also report numbers on sequences of length 2,4,5 and 7 in order to understand the performance as a function of scene length. We evaluate the methods on Sun3D dataset, in order to understand the generalization of our approach to other indoor scenes. We also discuss the multiply-accumuate operations (MACs) for the different methods to understand the operating efficiency at run-time. 

\subsection{Detector and Descriptor Quality}
Table \ref{table1} shows the results of the our detector and descriptor evaluation. Note that MLE and repeatability are detector metrics, MScore is a descriptor metric, and rotation@$5^{\circ}$  and translation@5cm are combined metrics. We set the threshold for our detector at 0.0005, the same as that used during training. This results in a large number of interest points being detected (Num) which artificially inflates the repeatability score (Rep) in our favour, but has poor localization performance as indicated by MLE metric. However, our MScore is comparable to SuperPoint although we trained our network to only match along the epipolar line, and not for the full image. Furthermore, we have the best rotation@$5^{\circ}$ and translation@5cm metric indicating that the matches found using our descriptors help accurately determine rotation and translation, i.e., pose. These results are indicative that our training procedure can complement the homographic adaptation technique of SuperPoint and boost the overall performance. Incorporation of evaluated pose using ideas discussed in \cite{sarlin2019superglue}, in lieu of ground truth pose to train our network is left for future work.   
\begin{table}
  \caption{Performance of different descriptors on ScanNet.}
  \centering
  \begin{tabular}{ccccccccc}
  \hline
     &\textbf{MLE}  &\textbf{MScore} & \textbf{Num} & \textbf{Rep}& \textbf{rot@$5^{\circ}$}& \textbf{trans@5cm}\\

    \hline
ORB  &	2.584 &		0.194  &   401  & 0.613 &  0.142 &		0.064	 \\
SIFT &	\textbf{2.327} &		0.201  &   203  & 0.496 &  0.311 &		0.148	\\
SURF &	2.577 &		0.198  &   268  & 0.460 &  0.303 &		0.134	\\
SuperPoint  &  2.545 &     \textbf{0.375}  &   129  & 0.519	&  0.489 &      0.244 \\
Ours &	 3.101 &   0.329   &  1511  & \textbf{0.738} &  \textbf{0.518} &\textbf{0.254} 	\\
\hline
  \end{tabular}
  \label{table1}
\end{table}

\subsection{Depth Results}
\begin{figure}
\centering
\includegraphics[width=1.0\textwidth]{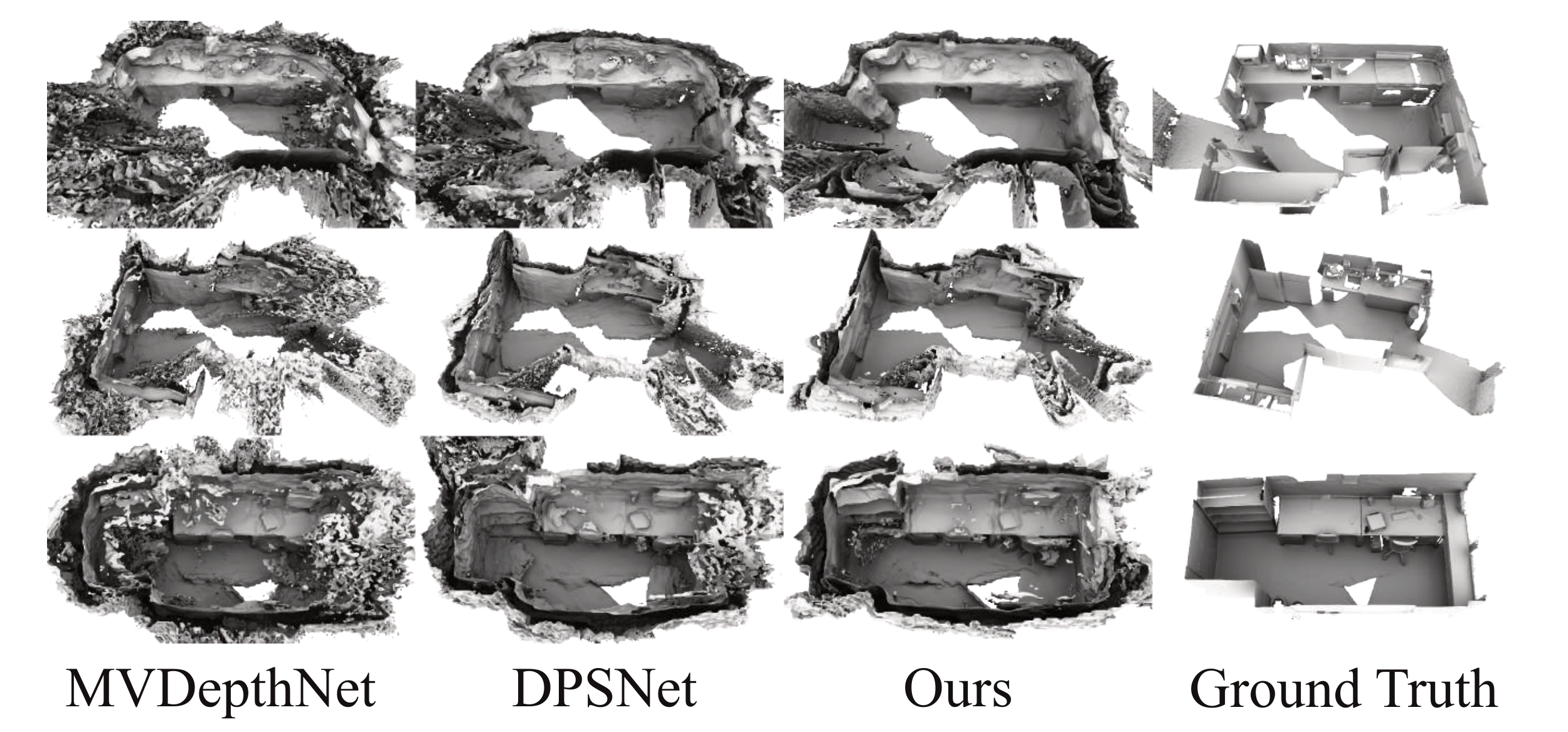}
\caption{3D scene reconstruction using predicted depth over the full sequence.} 
\label{geo}
\end{figure}

\begin{figure}[htp]
  \centering
    \begin{tabular}{cccccc}
        \includegraphics[width=.16\textwidth,height=1.5cm]{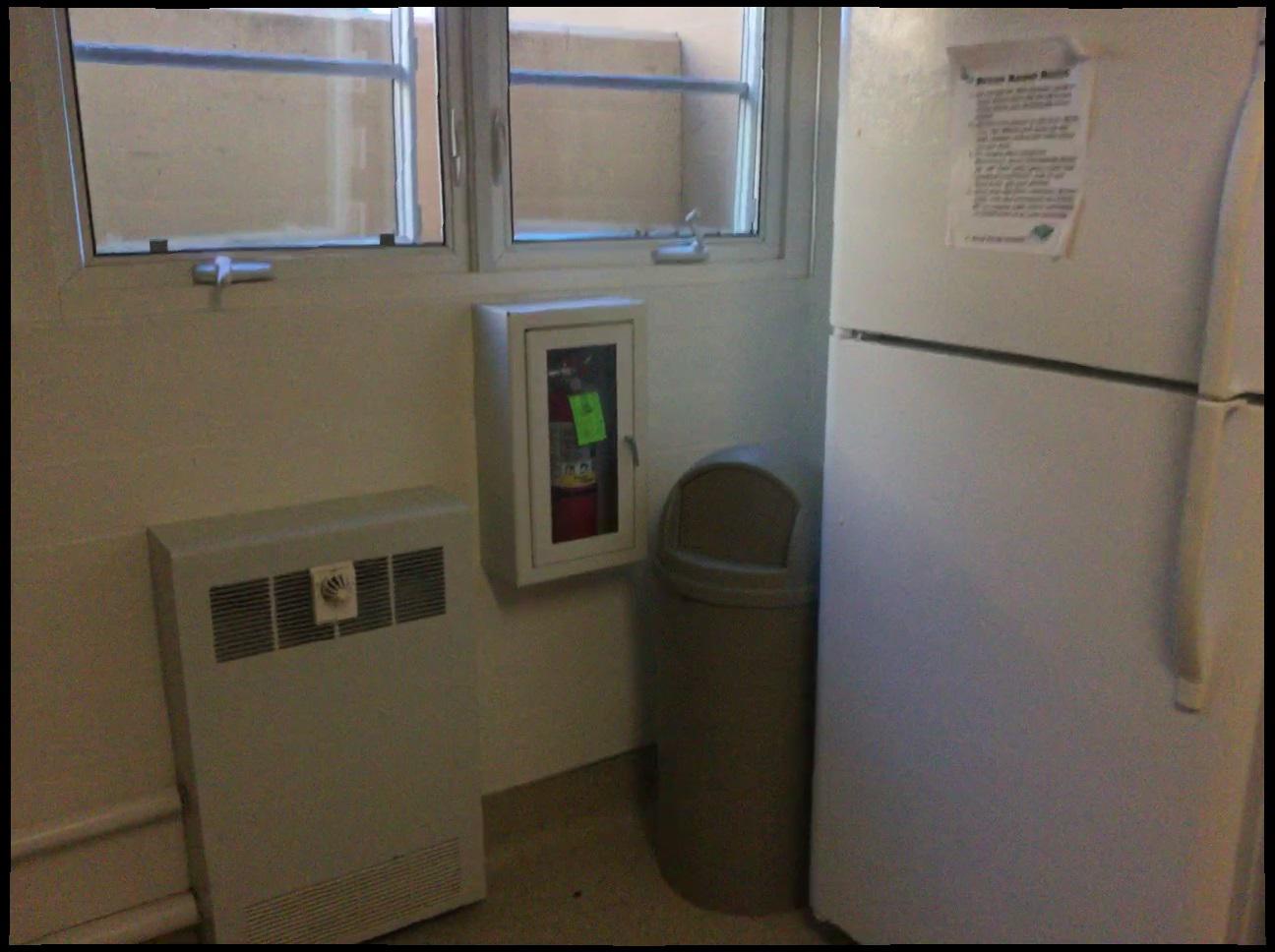} &
        \includegraphics[width=.16\textwidth,height=1.5cm]{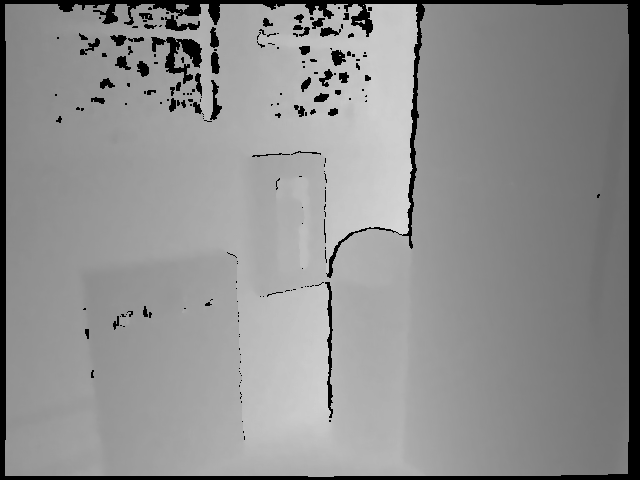} &    \includegraphics[width=.16\textwidth,height=1.5cm]{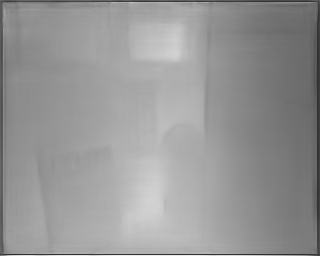} &    \includegraphics[width=.16\textwidth,height=1.5cm]{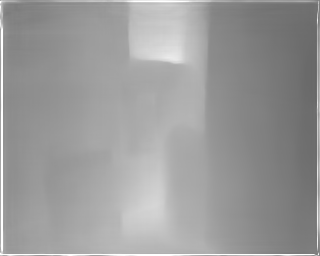} &    \includegraphics[width=.16\textwidth,height=1.5cm]{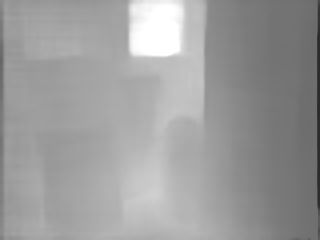} &    \includegraphics[width=.16\textwidth,height=1.5cm]{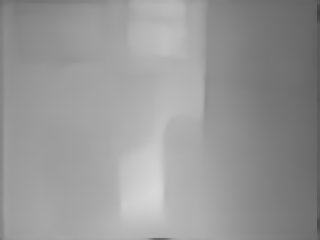} \\
                \includegraphics[width=.16\textwidth,height=1.5cm]{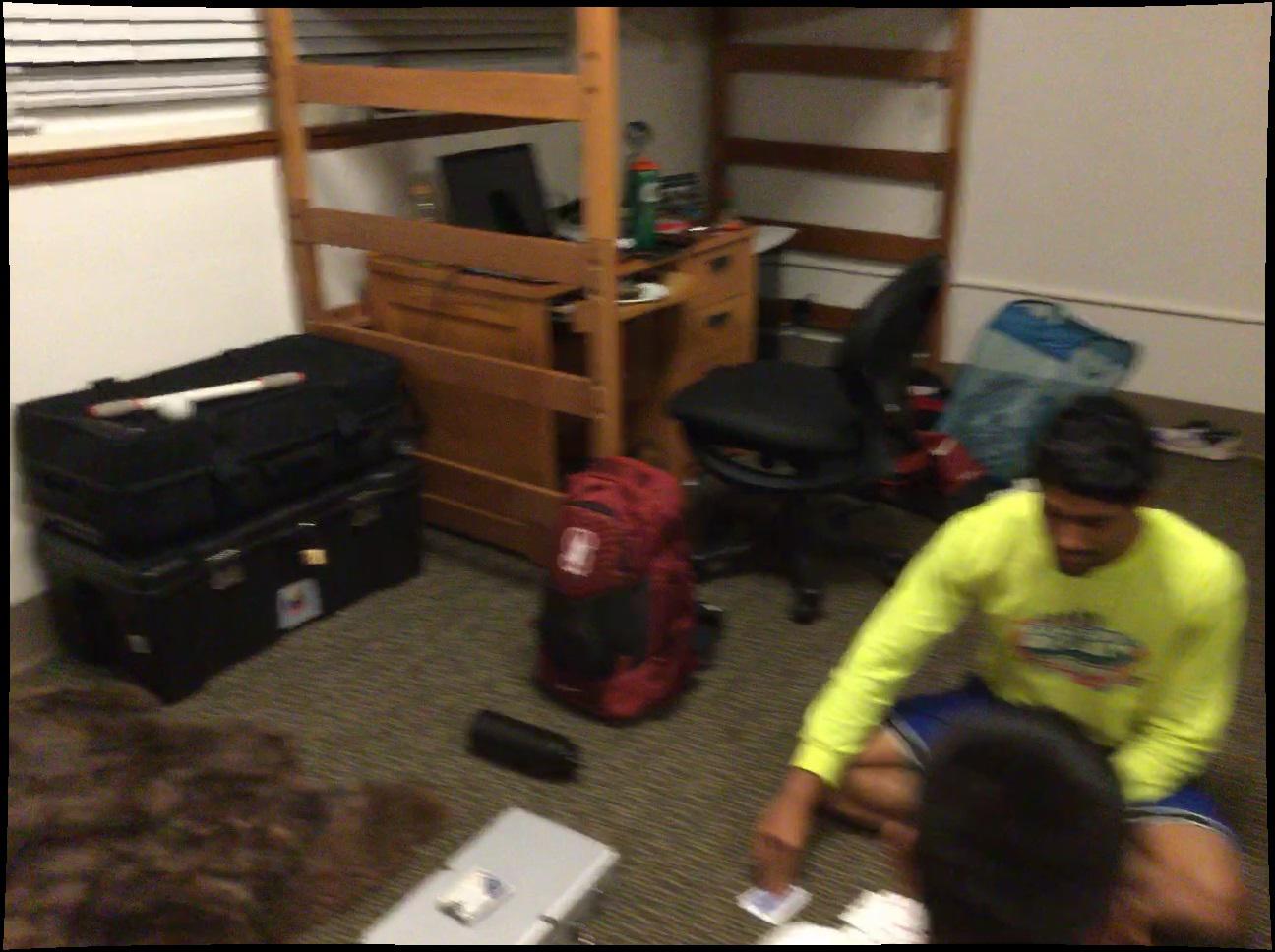} &
        \includegraphics[width=.16\textwidth,height=1.5cm]{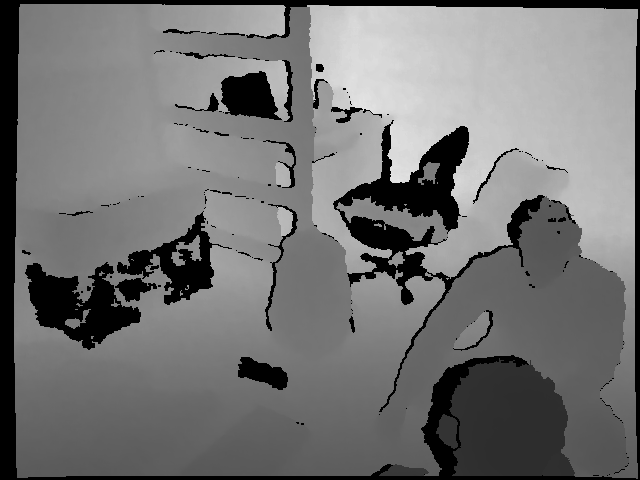} &    \includegraphics[width=.16\textwidth,height=1.5cm]{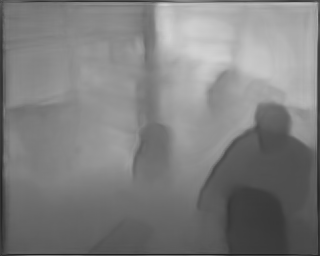} &    \includegraphics[width=.16\textwidth,height=1.5cm]{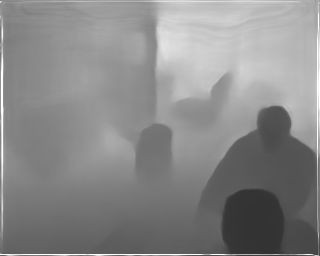} &    \includegraphics[width=.16\textwidth,height=1.5cm]{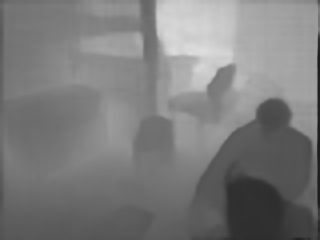} &    \includegraphics[width=.16\textwidth,height=1.5cm]{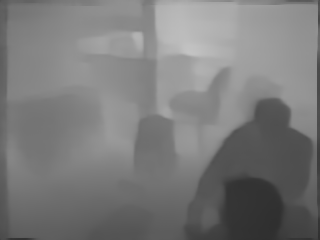} \\         \includegraphics[width=.16\textwidth,height=1.5cm]{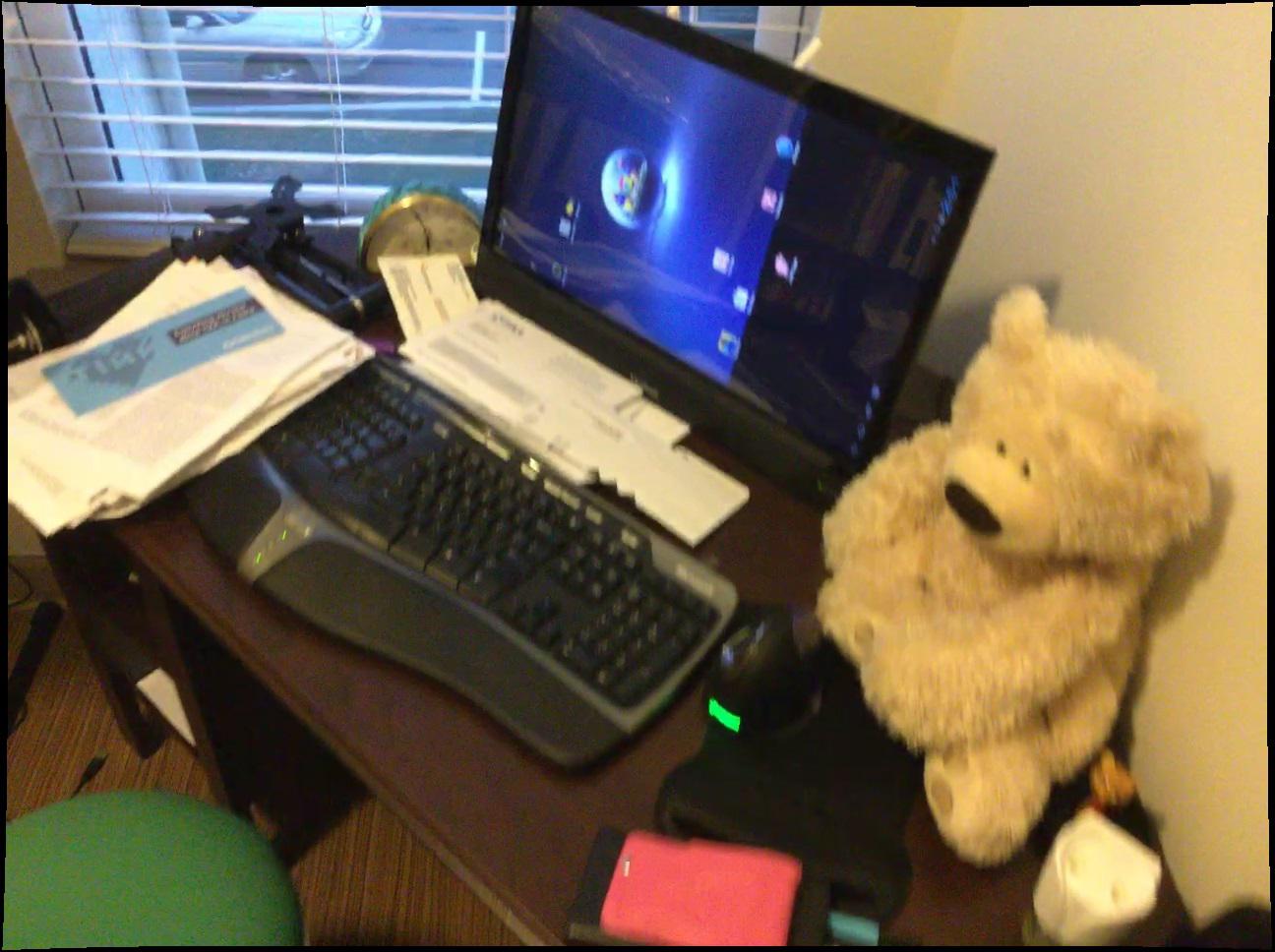} &
        \includegraphics[width=.16\textwidth,height=1.5cm]{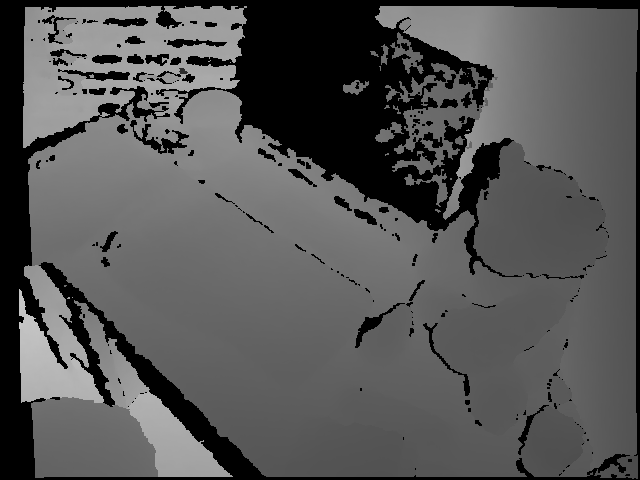} &    \includegraphics[width=.16\textwidth,height=1.5cm]{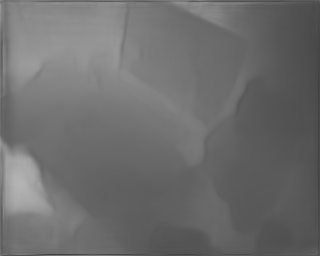} &    \includegraphics[width=.16\textwidth,height=1.5cm]{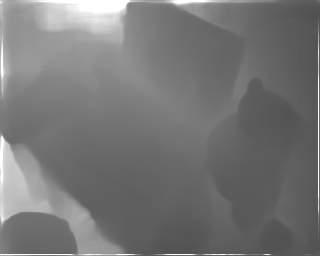} &    \includegraphics[width=.16\textwidth,height=1.5cm]{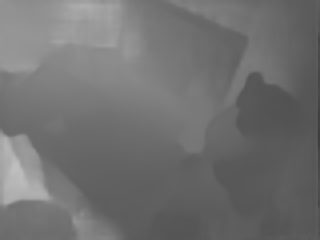} &    \includegraphics[width=.16\textwidth,height=1.5cm]{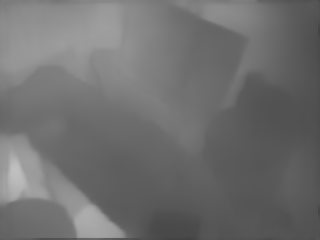} \\         \includegraphics[width=.16\textwidth,height=1.5cm]{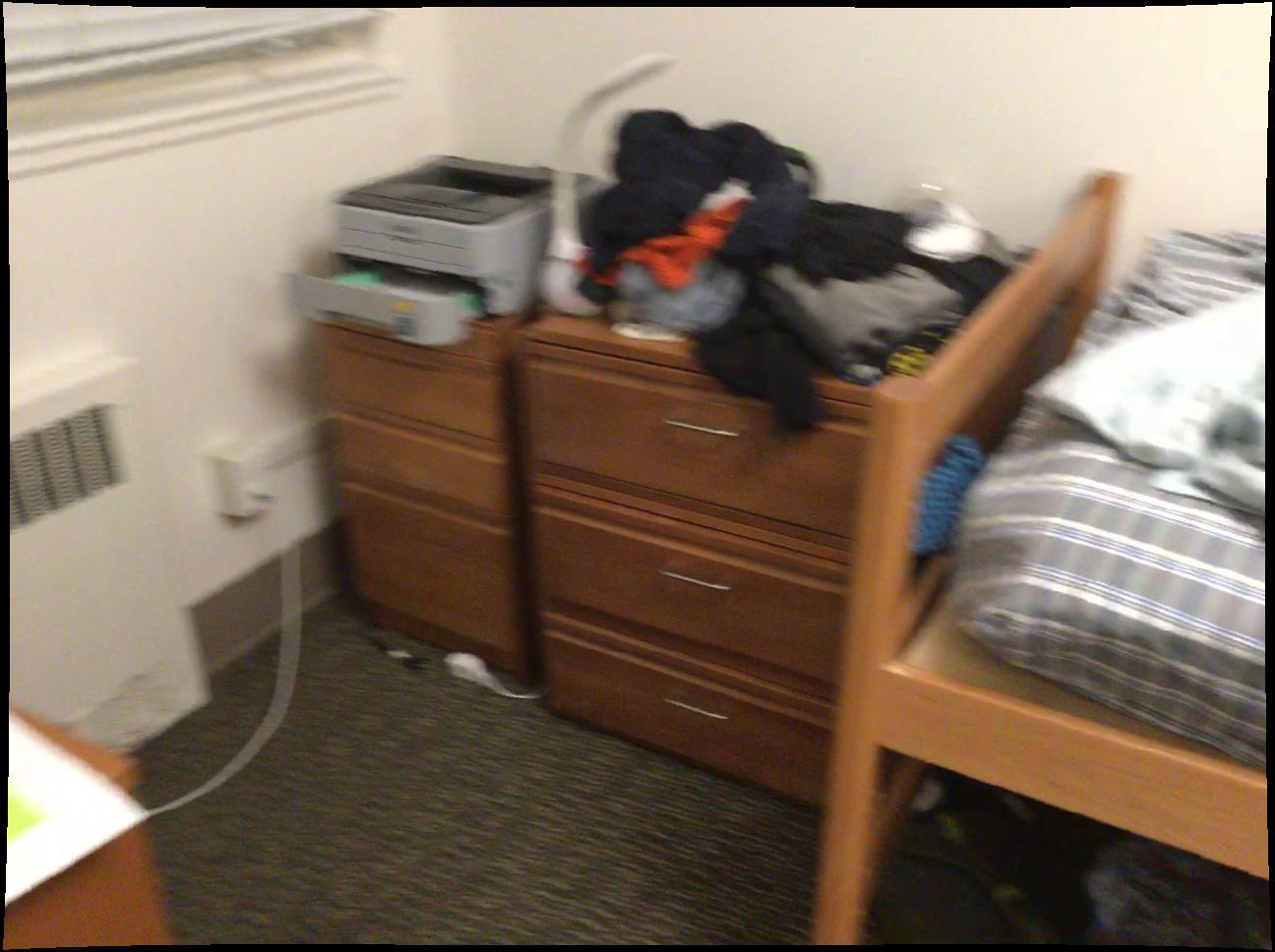} &
        \includegraphics[width=.16\textwidth,height=1.5cm]{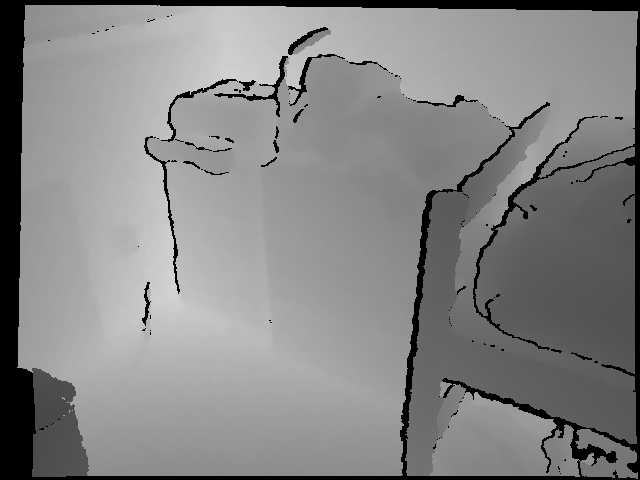} &    \includegraphics[width=.16\textwidth,height=1.5cm]{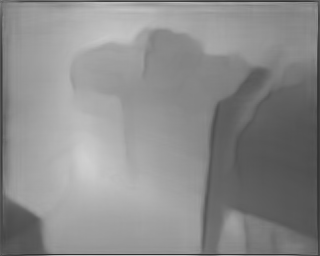} &    \includegraphics[width=.16\textwidth,height=1.5cm]{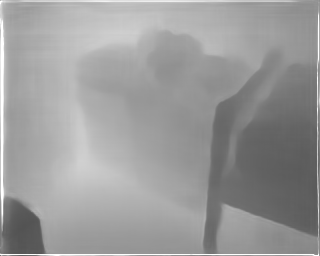} &    \includegraphics[width=.16\textwidth,height=1.5cm]{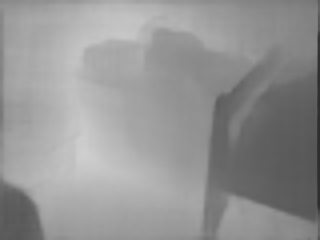} &    \includegraphics[width=.16\textwidth,height=1.5cm]{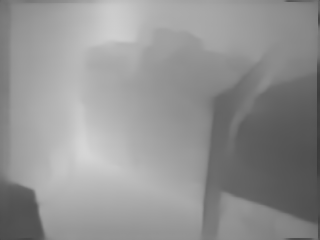} \\         \includegraphics[width=.16\textwidth,height=1.5cm]{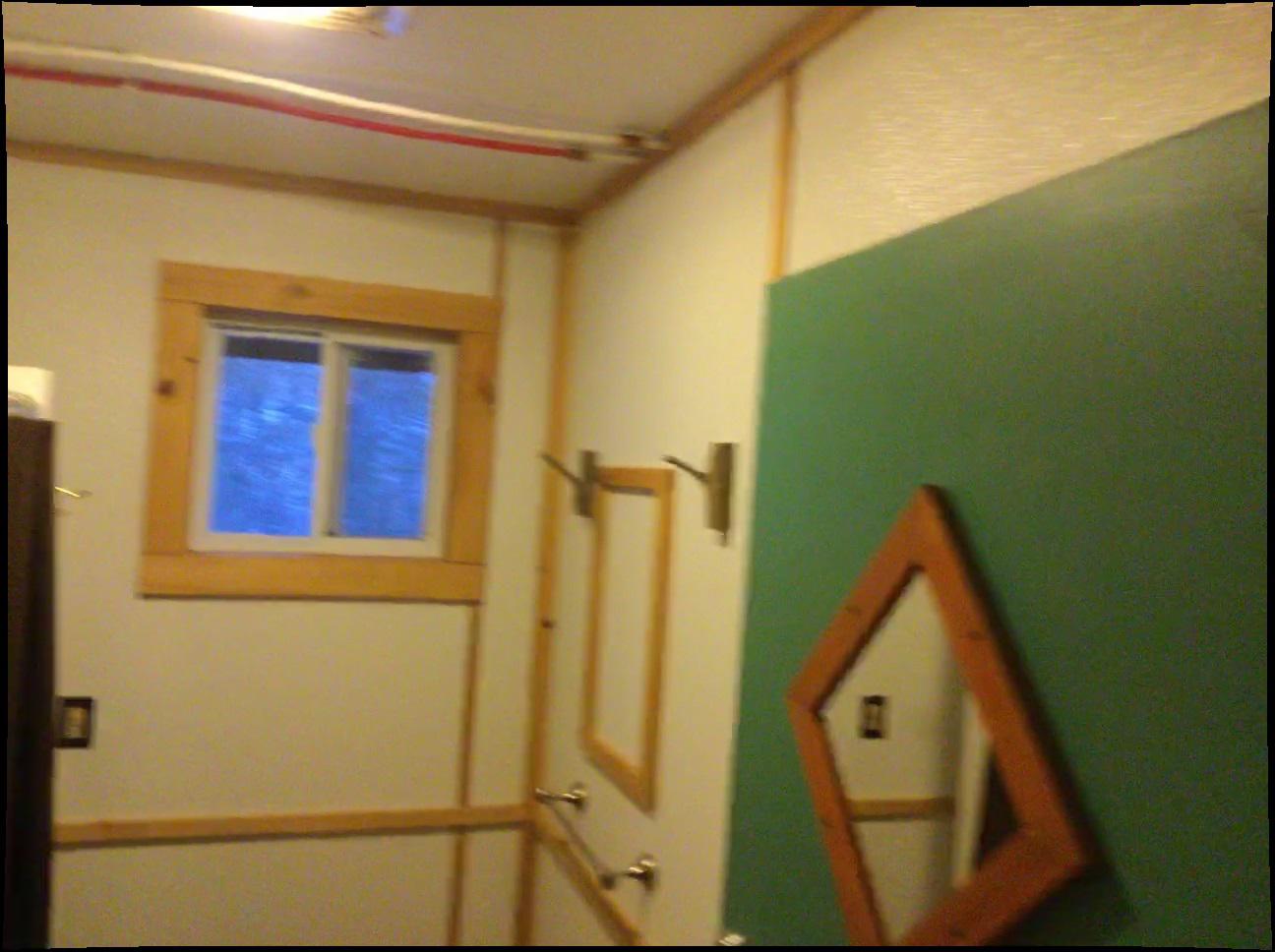} &
        \includegraphics[width=.16\textwidth,height=1.5cm]{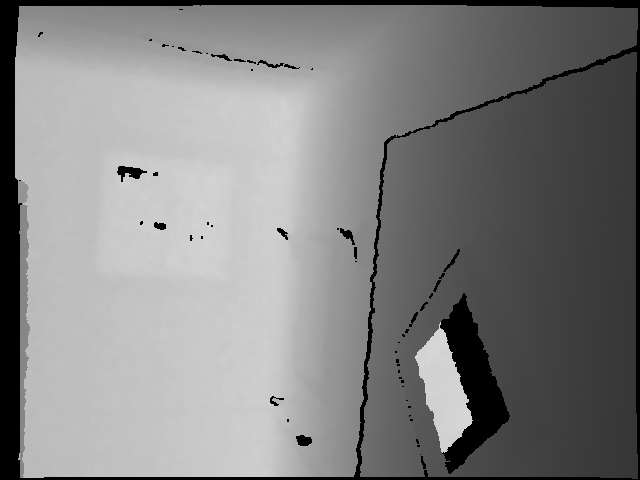} &    \includegraphics[width=.16\textwidth,height=1.5cm]{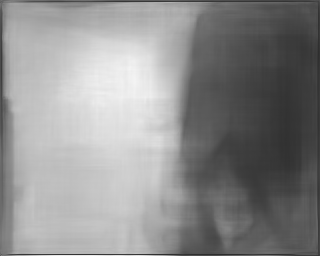} &    \includegraphics[width=.16\textwidth,height=1.5cm]{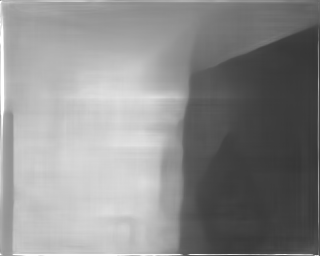} &    \includegraphics[width=.16\textwidth,height=1.5cm]{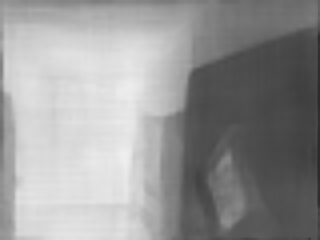} &    \includegraphics[width=.16\textwidth,height=1.5cm]{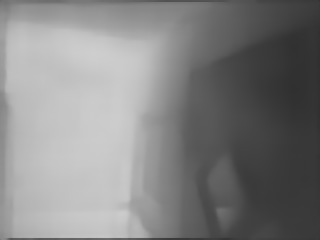} \\         \includegraphics[width=.16\textwidth,height=1.5cm]{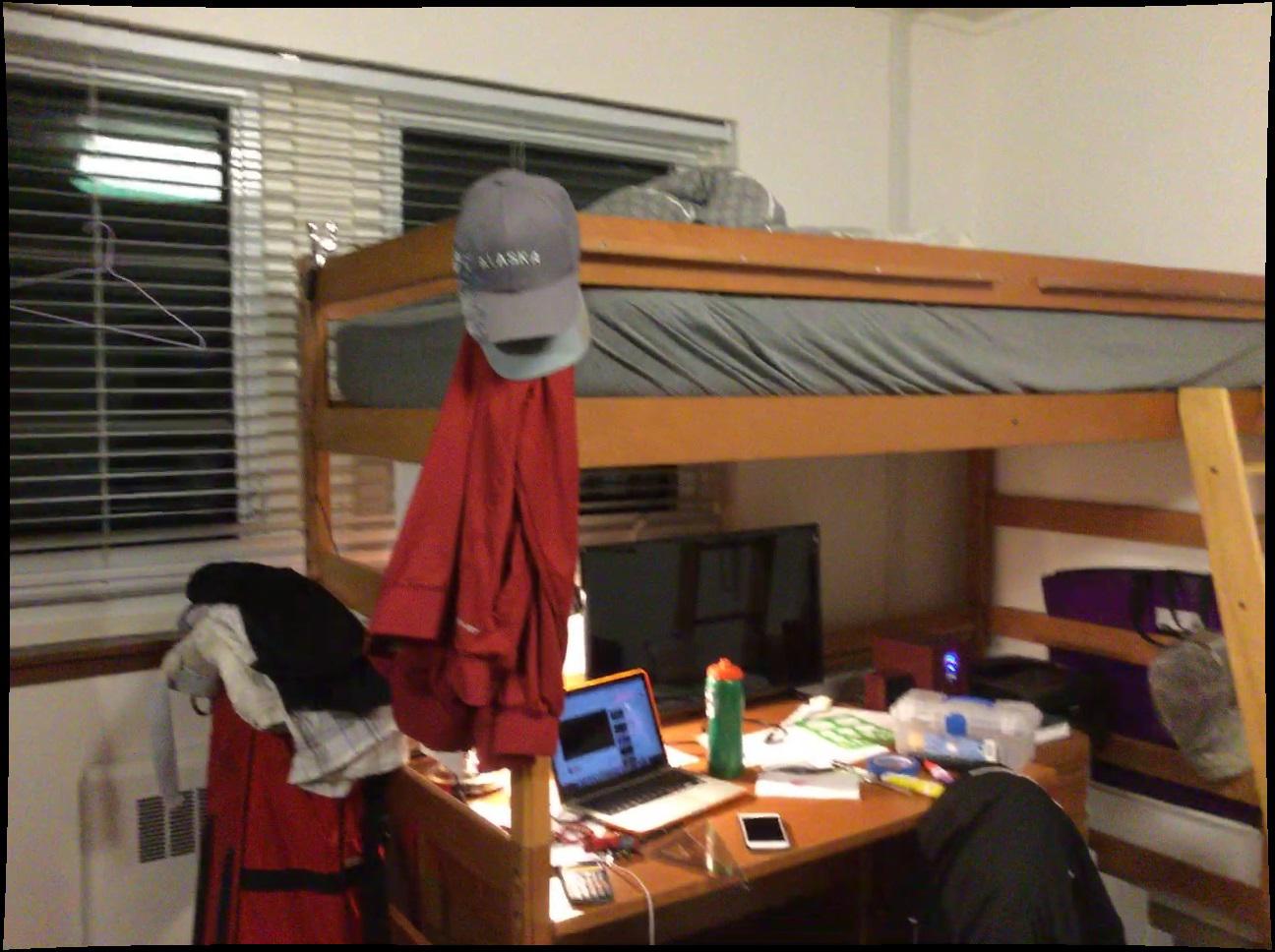} &
        \includegraphics[width=.16\textwidth,height=1.5cm]{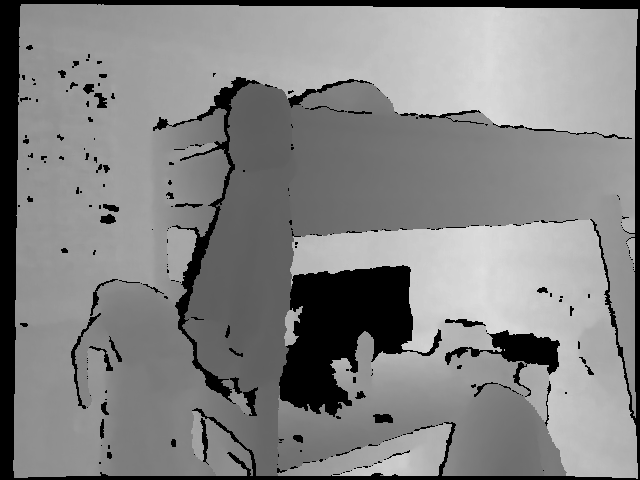} &    \includegraphics[width=.16\textwidth,height=1.5cm]{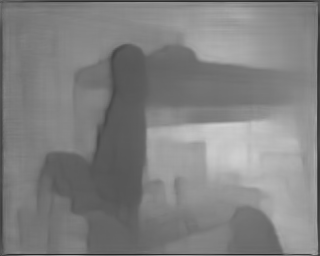} &    \includegraphics[width=.16\textwidth,height=1.5cm]{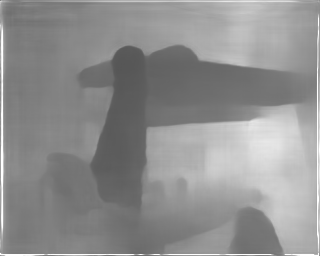} &    \includegraphics[width=.16\textwidth,height=1.5cm]{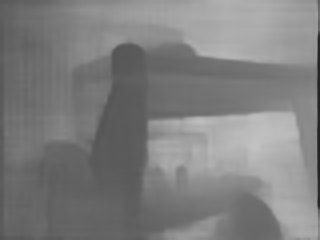} &    \includegraphics[width=.16\textwidth,height=1.5cm]{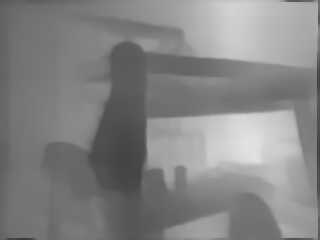} \\         \includegraphics[width=.16\textwidth,height=1.5cm]{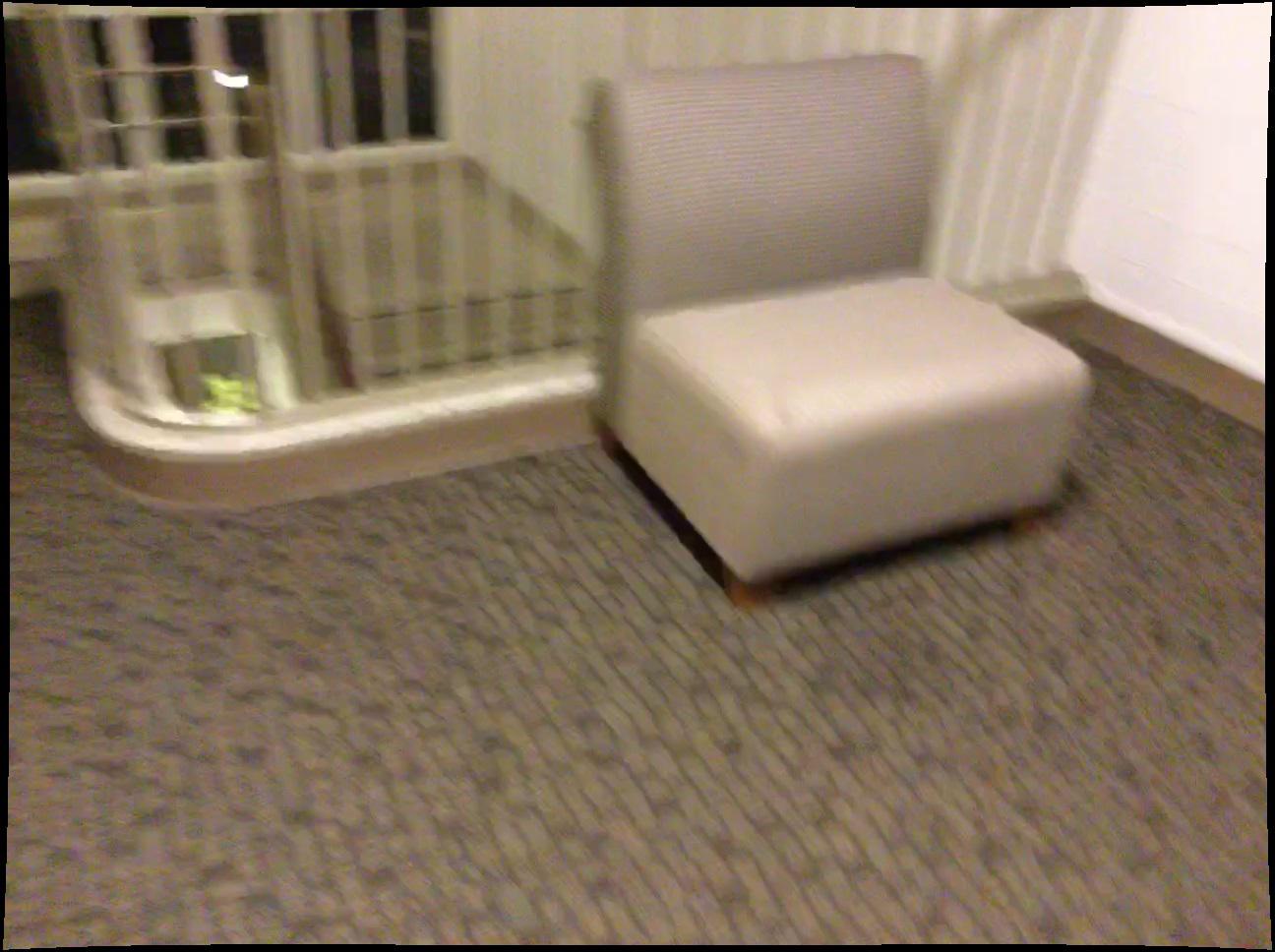} &
        \includegraphics[width=.16\textwidth,height=1.5cm]{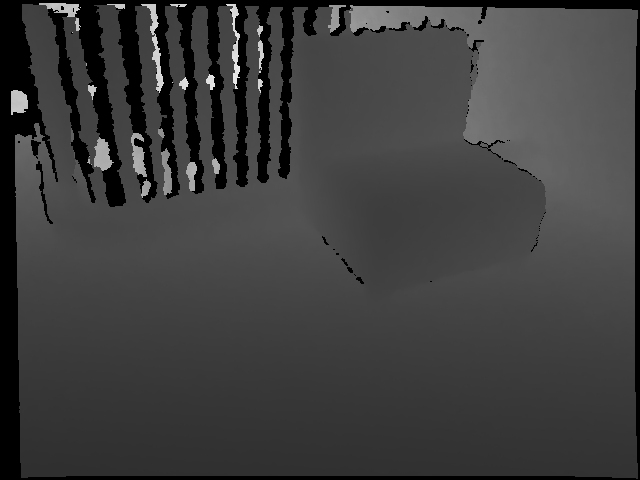} &    \includegraphics[width=.16\textwidth,height=1.5cm]{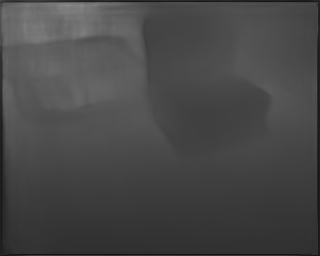} &    \includegraphics[width=.16\textwidth,height=1.5cm]{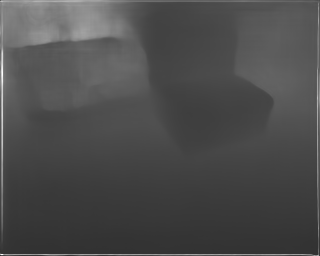} &    \includegraphics[width=.16\textwidth,height=1.5cm]{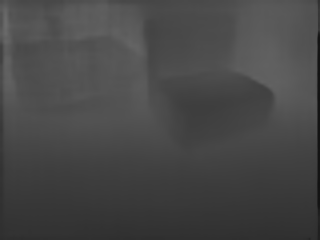} &    \includegraphics[width=.16\textwidth,height=1.5cm]{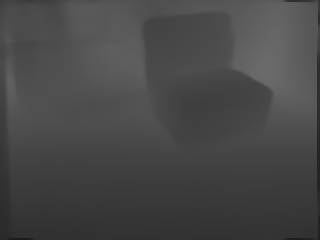} \\         \includegraphics[width=.16\textwidth,height=1.5cm]{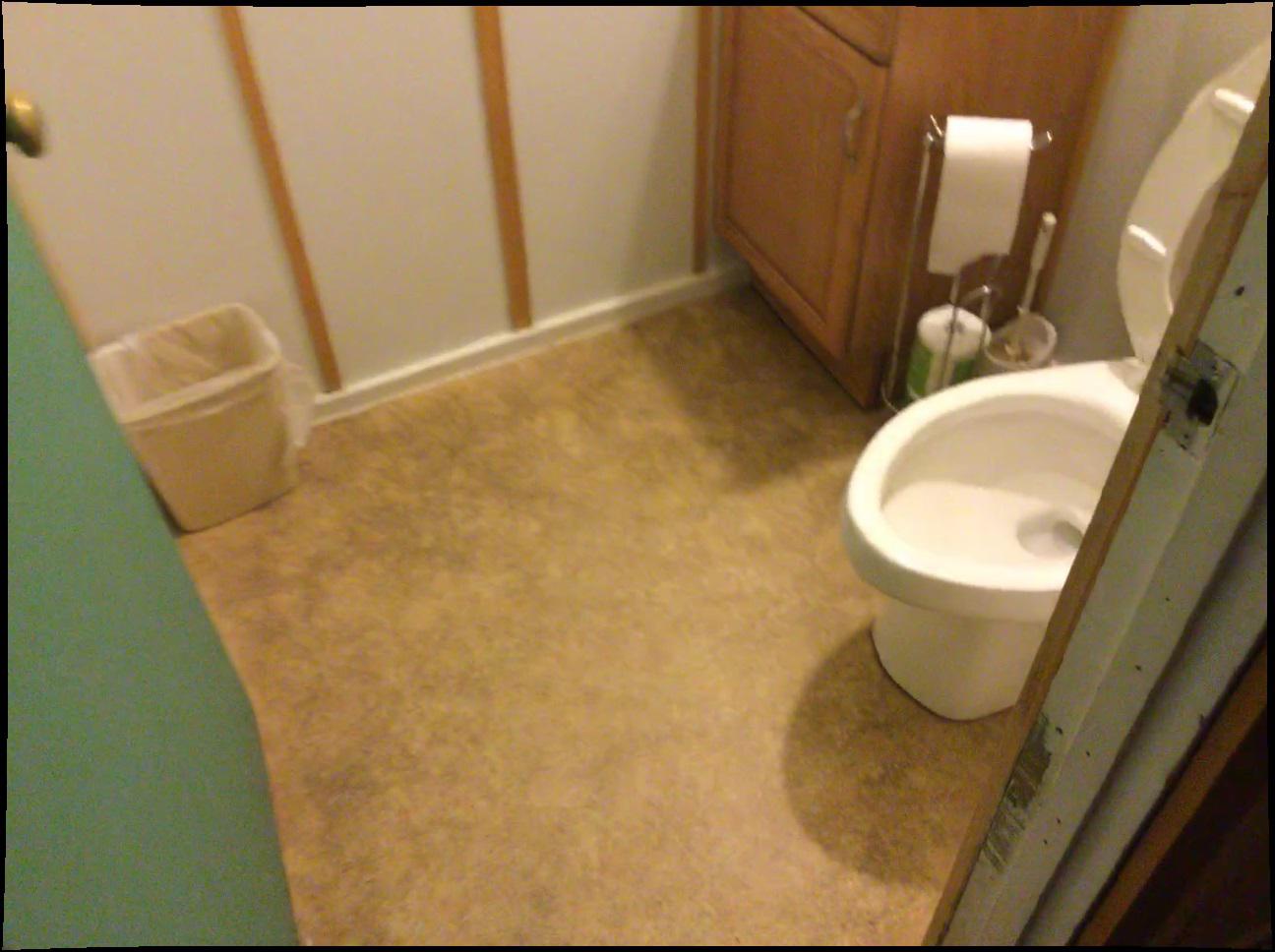} &
        \includegraphics[width=.16\textwidth,height=1.5cm]{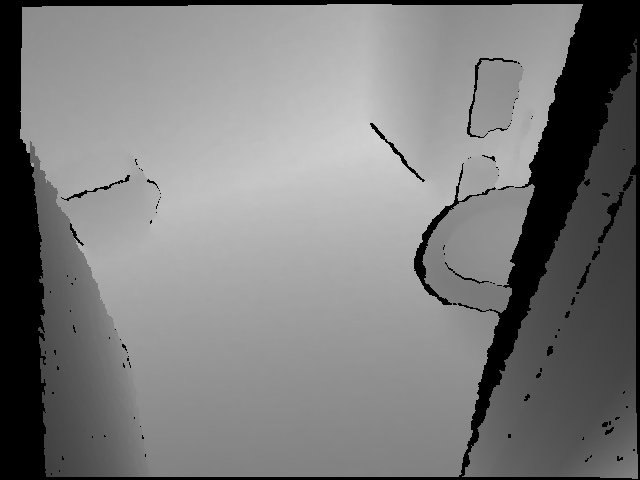} &    \includegraphics[width=.16\textwidth,height=1.5cm]{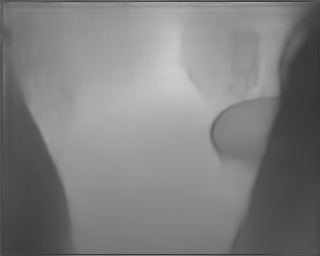} &    \includegraphics[width=.16\textwidth,height=1.5cm]{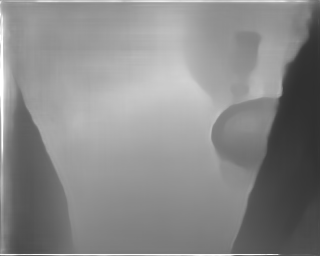} &    \includegraphics[width=.16\textwidth,height=1.5cm]{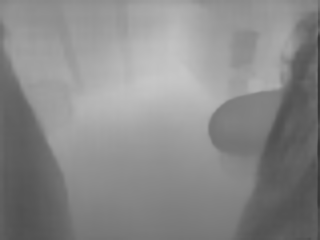} &    \includegraphics[width=.16\textwidth,height=1.5cm]{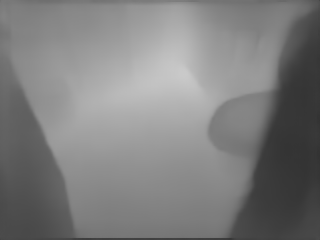} \\         \includegraphics[width=.16\textwidth,height=1.5cm]{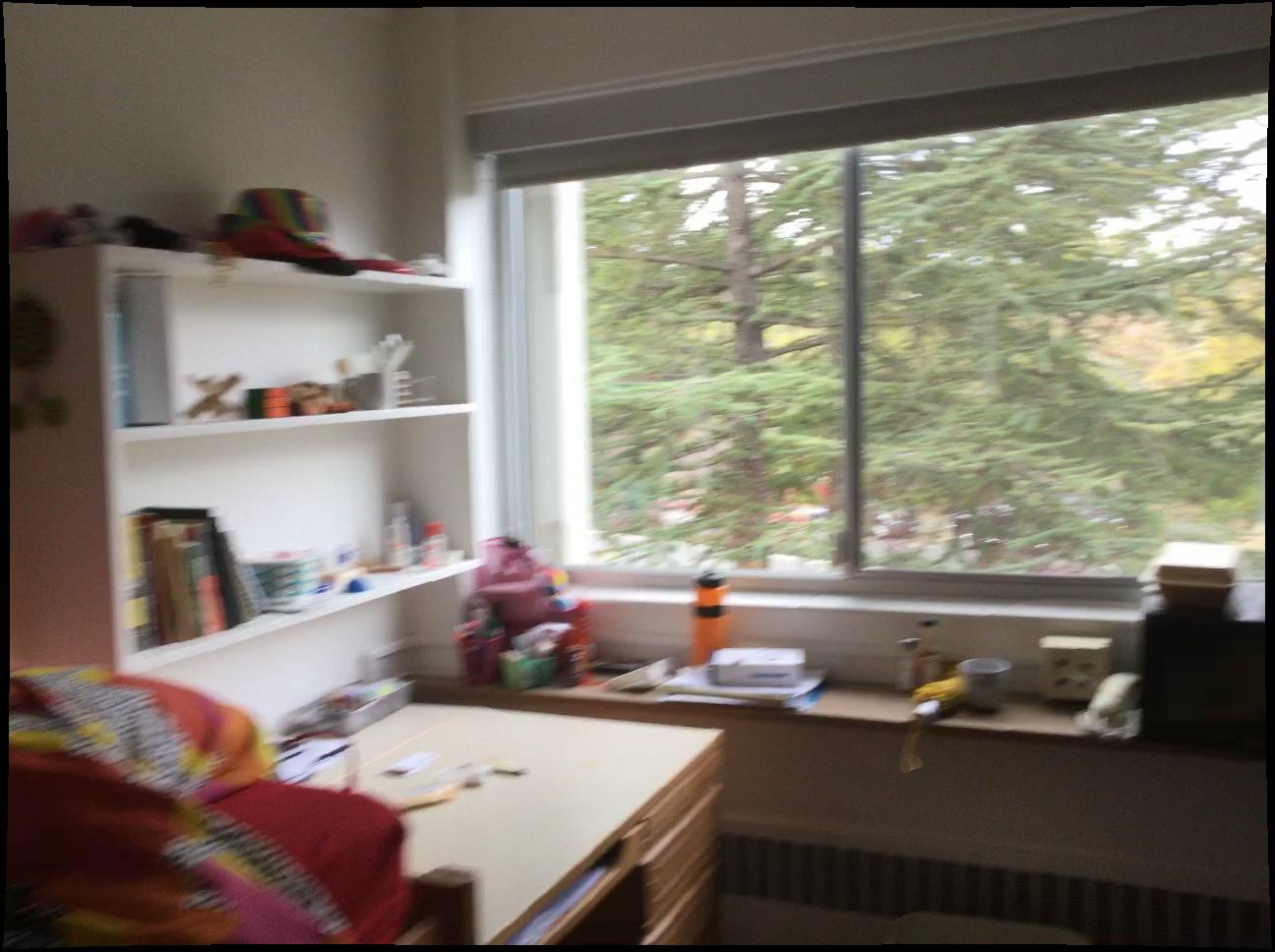} &
        \includegraphics[width=.16\textwidth,height=1.5cm]{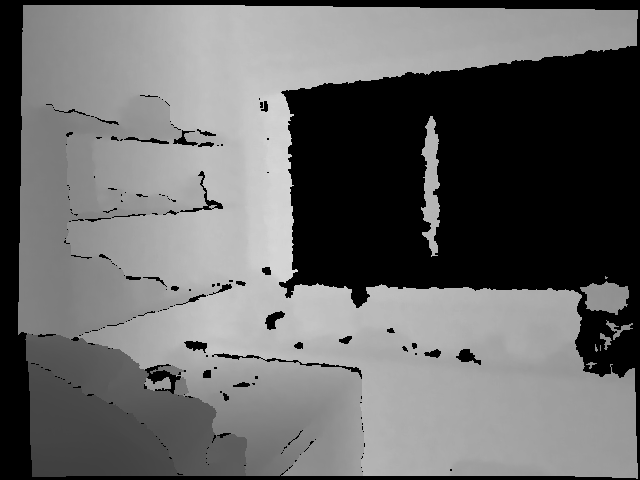} &    \includegraphics[width=.16\textwidth,height=1.5cm]{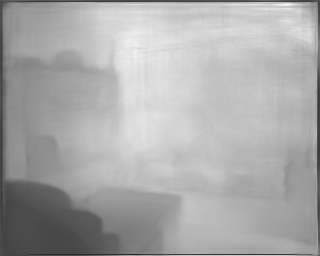} &    \includegraphics[width=.16\textwidth,height=1.5cm]{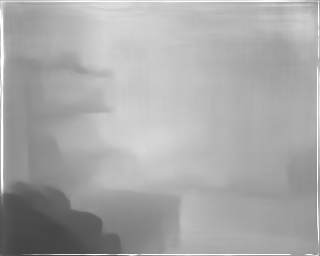} &    \includegraphics[width=.16\textwidth,height=1.5cm]{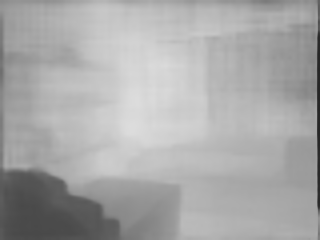} &    \includegraphics[width=.16\textwidth,height=1.5cm]{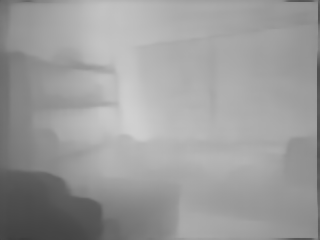} \\         \includegraphics[width=.16\textwidth,height=1.5cm]{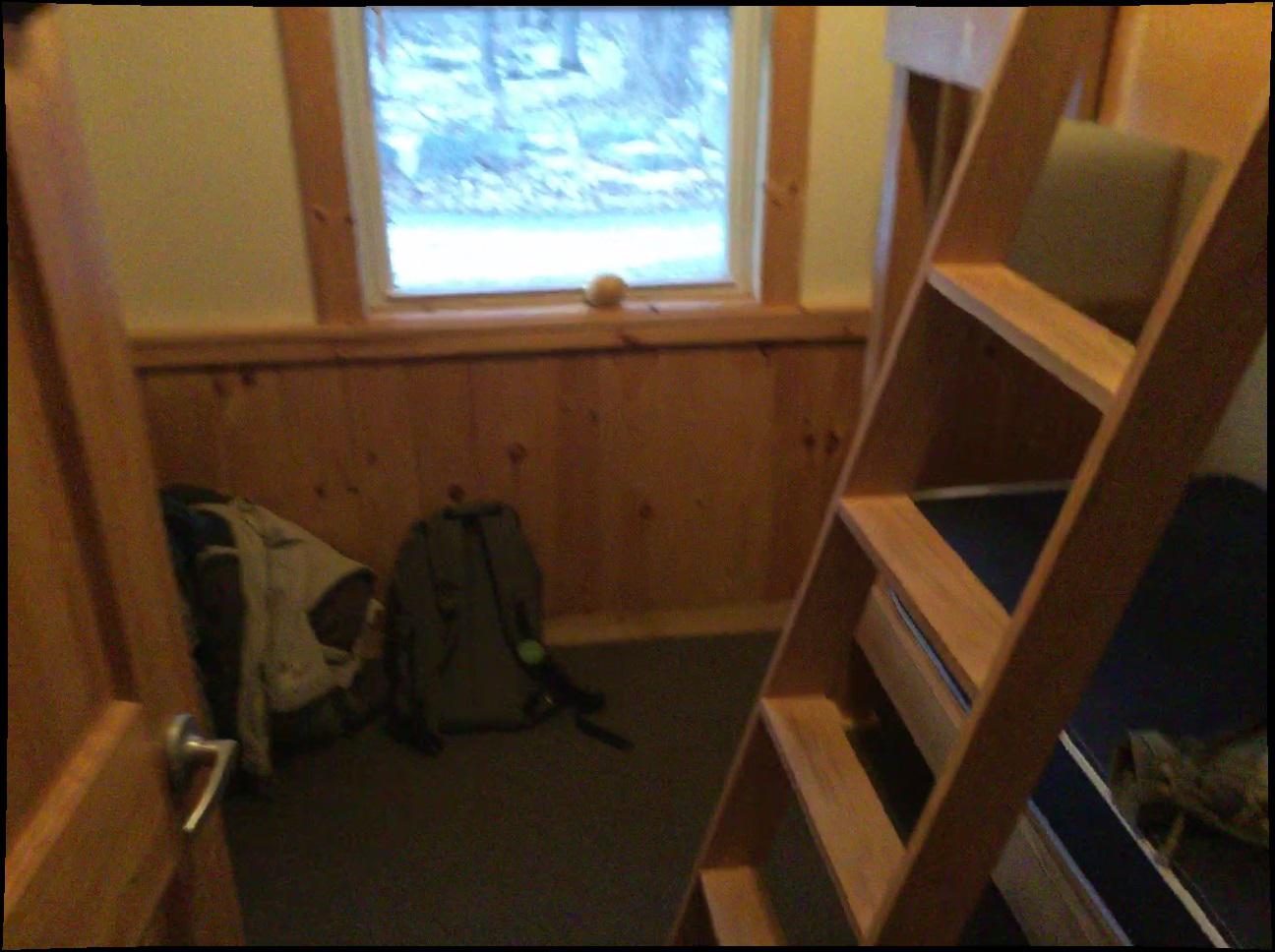} &
        \includegraphics[width=.16\textwidth,height=1.5cm]{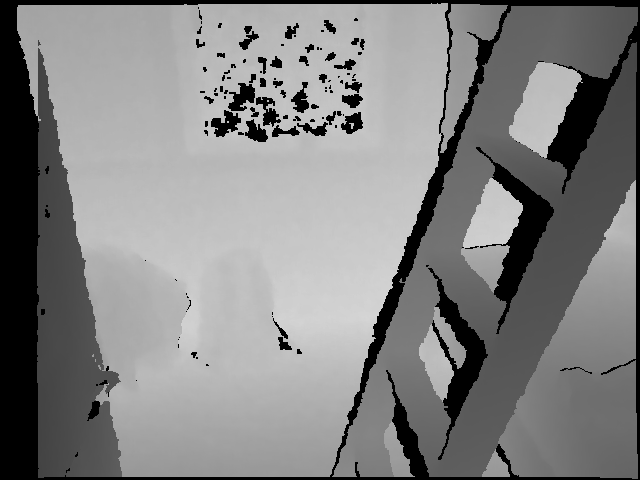} &    \includegraphics[width=.16\textwidth,height=1.5cm]{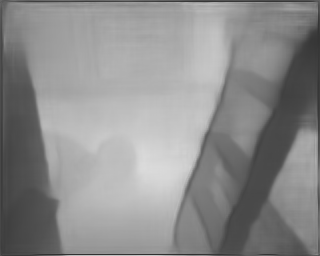} &    \includegraphics[width=.16\textwidth,height=1.5cm]{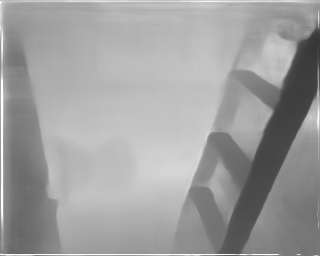} &    \includegraphics[width=.16\textwidth,height=1.5cm]{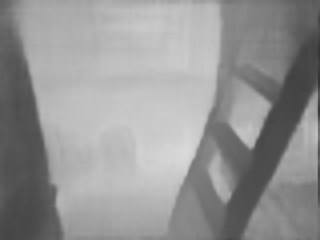} &    \includegraphics[width=.16\textwidth,height=1.5cm]{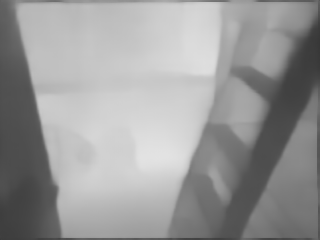} \\         \includegraphics[width=.16\textwidth,height=1.5cm]{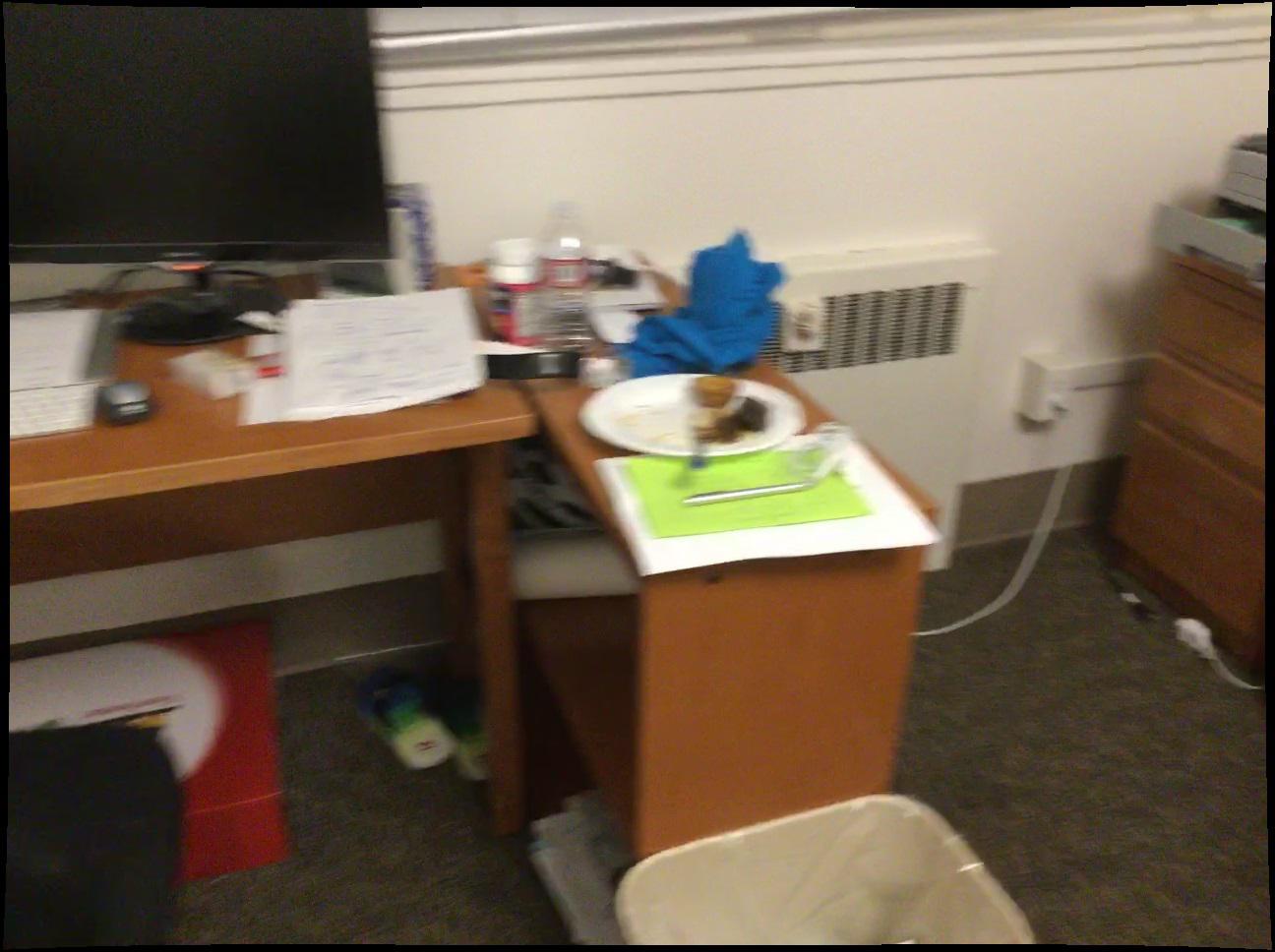} &
        \includegraphics[width=.16\textwidth,height=1.5cm]{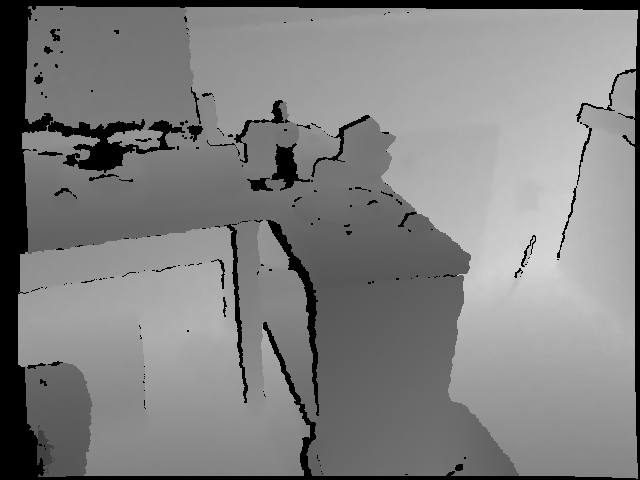} &    \includegraphics[width=.16\textwidth,height=1.5cm]{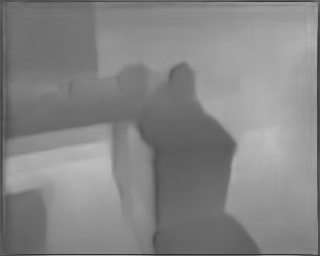} &    \includegraphics[width=.16\textwidth,height=1.5cm]{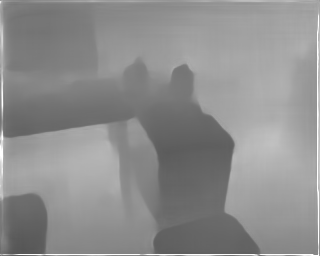} &    \includegraphics[width=.16\textwidth,height=1.5cm]{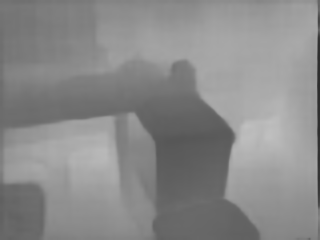} &    \includegraphics[width=.16\textwidth,height=1.5cm]{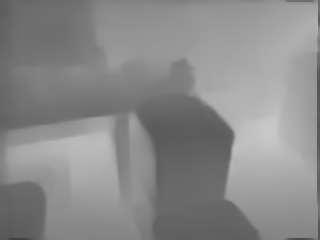} \\         
        Image & GT Depth& MVDepthNet & GPMVSNet & DPSNet & Ours \\
    \end{tabular}
    \caption{Qualitative Performance of our networks on sampled images from ScanNet.}
  \label{fig:device2}
 \end{figure}

We set the same hyper-parameters for evaluating our network for all scenarios and across all datasets, i.e., fix the number of points detected to be 512, length of the sampled descriptors to be 100, and the detector threshold to be 5e-4. In order to ensure uniform distribution of the interest points and avoid clusters, we set a high NMS value of 9 as suggested in \cite{superpoint}. The supplement has analysis of the sparse depth output from our network and ablation study over different choices of hyper parameters. Table \ref{table2} shows the performance of depth estimation on sequences of length 3 and gap 20 as used in the training set. For fair comparison, we evaluate two versions of the competing approaches (1) The author provided open source trained model, (2) The trained model fine-tuned on ScanNet for 100K iterations with the default training parameters as suggested in the manuscript or made available by the authors. We use a gap of 20 frames to train each network, similar to ours. The fine-tuned models are indicated by the suffix \emph{FT} in the table. Unsurprisingly, the fine-tuned models fare much better than the original models on ScanNet evaluation. MVDepthNet has least improvement after fine-tuning, which can be attributed to the heavy geometric and photometric augmentation used during training, hence making it generalize well. DPSNet benefits maximally from fine-tuning with over 25\% drop in absolute error. However, our network outperforms all methods across all metrics. Figure \ref{fig:device2} shows qualitative comparison between the different methods and Figure \ref{geo} show sample 3D reconstructions of the scene from the estimated depth maps. In Figure \ref{fig:device2}, we see that MVDepthNet has gridding artifacts, which are removed by GPMVSNet. However, GPMVSNet has poor metric performance. DPSNet washes away finer details and also suffers from gridding artifacts. Our method preserves finer details while maintaining global coherence compared to all other methods. As we use geometry to estimate sparse depth, and the network in-fills the missing values, we retain metric performance while leveraging the generative ability of CNNs with sparse priors. In Figure \ref{geo} we see our method consistently output less noisy scene reconstructions compared to MVDepthNet and DPSNet. Moreover, we see planes and corners being respected better than the other methods. 
\begin{table}[t]
  \caption{Performance of depth estimation on ScanNet. We use sequences of length 3 and sample every 20 frames. FT indicates fine-tuned on ScanNet. }
  \centering
  \begin{tabular}{ccccccccccc}
 \hline 
     &Abs Rel  &Abs & Sq Rel & RMSE & RMSE log & \textbf{$\delta< 1.25$}& \textbf{$\delta < 1.25^2$}& \textbf{$\delta< 1.25^3$}\\

    \hline
GPMVS &	0.1306 &0.2600  &  0.0944  & 0.3451 & 0.1881 & 0.8481 &  0.9462 &	0.9753	\\
GPMVS-FT  &  0.1079 & 0.2255 & 0.0960 & 0.4659 & 0.1998 & 0.8905 & 0.9591 & 0.9789 \\
MVDepth & 0.1191 & 0.2096 & 0.0910 & 0.3048 & 0.1597 & 0.8690 & 0.9599 & 0.9851 \\
MVDepth-FT & 0.1054 & 0.1911 & 0.0970 & 0.3053 & 0.1553& 0.8952 & 0.9707 & 0.9895  \\
DPS & 0.1470 & 0.2248 & 0.1035 & 0.3468& 0.1952 & 0.8486 & 0.9474 & 0.9761		 \\
DPS-FT &	0.1025 & 0.1675 & 0.0574 & 0.2679 & 0.1531 & 0.9102 & 0.9708 & 0.9872	\\
Ours &	\textbf{0.0932} & \textbf{0.1540} & \textbf{0.0506} & \textbf{0.2505} & \textbf{0.1426}& \textbf{0.9287} & \textbf{0.9767} & \textbf{0.9893} \\
\hline
  \end{tabular}
\label{table2}
\end{table}

An important feature of any multiview stereo method is the ability to improve with more views. Table \ref{table3} shows the performance for different number of images. We set the frame gap to be 20, 15, 12 and 10 for 2,4,5 and 7 frames respectively. These gaps ensure that each set approximately span similar volumes in 3D space, and any performance improvement emerges from the network better using the available information as opposed to acquiring new information. We again see that our method outperforms all other methods on all three metrics for different sequence lengths. Closer inspection of the values indicate that the DPSNet and GPMVSNet do not benefit from additional views, whereas, MVDepthNet benefits from a small number of additional views but stagnates for more than 4 frames. On the contrary, we show steady improvement in all three metrics with additional views. This can be attributed to our point matcher and triangulation module which naturally benefits from additional views. 
\begin{table}[t]
  \caption{Performance of depth estimation on ScanNet. Results on sequences of various lengths are presented. GPN: GPMVSNet, MVN: MVDepthNet, DPS: DPSNet. AbR: Abolute Relative, Abs: Absolute difference, SqR: Square Relative.}
  \centering
  \begin{tabular}{cccccccccccccc}
 \hline 
   Method    & \multicolumn{3}{c}{\textbf{2 Frames} } & \multicolumn{3}{c}{\textbf{4 Frames}}& \multicolumn{3}{c}{\textbf{5 Frames}}& \multicolumn{3}{c}{\textbf{7 Frames}}\\ \cline{2-13}
     &AbR  &Abs & SqR  &AbR  &Abs & SqR &AbR  &Abs & SqR &AbR  &Abs & SqR\\

    \hline
GPN & 0.112 & 0.233 & 0.101 & 0.109 & 0.226 & 0.100 & 0.107 & 0.226 & 0.112 & 0.109 & 0.230 & 0.116\\

MVN &  0.126 & 0.238 & 0.471 & 0.105 & 0.191 & 0.078 & 0.106 & 0.192 & 0.071 & 0.108 & 0.195 & 0.067 \\

DPS & \textbf{0.099} & 0.181 & 0.062 & 0.102 & 0.168 & 0.057 & 0.102 & 0.168 & 0.057 & 0.102 & 0.167 & 0.057		 \\

Ours & 	0.106 & \textbf{0.173} & \textbf{0.057} & \textbf{0.090} & \textbf{0.150} & \textbf{0.049} & \textbf{0.088} & \textbf{0.147} & \textbf{0.048} & \textbf{0.087} & \textbf{0.144} & \textbf{0.043} \\
\hline
  \end{tabular}
  \label{table3}
\end{table}

As a final experiment, we test our network on Sun3D test dataset consisting of 80 pairs of images. Sun3D also captures indoor environments, albeit at a much smaller scale compared to ScanNet. Table \ref{sun3d} shows the performance for the two versions of DPSNet and MVDepthNet discussed previously, and our network. Note DPSNet and MVDepthNet were originally trained on the Sun3D training database. The fine-tuned version of DPSNet performs better than the original network on the Sun3D test set owing to the greater diversity in ScanNet training database. MVDepthNet on the contrary performs worse, indicating that it overfit to ScanNet and the original network was sufficiently trained and generalized well. Remarkably, we again outperform both methods although our trained network has never seen any image from the Sun3D database. This indicates that our principled way of determining sparse depth, and then densifying has good generalizability. The supplement shows additional qualitative results. 
\begin{table}[t]
  \caption{Performance of depth estimation on Sun3D. We use sequences of length 2.}
  \centering
  \begin{tabular}{ccccccccccc}
    \hline
     &Abs Rel  &Abs & Sq Rel & RMSE & RMSE log & \textbf{$\delta< 1.25$}& \textbf{$\delta < 1.25^2$}& \textbf{$\delta< 1.25^3$}\\
    \hline
    MVDepth & 0.1377  &  0.3199 & 0.1564 & 0.4523 & 0.1853 & 0.8245 & 0.9601 & 0.9851  \\
    MVDepth-FT & 0.3092 & 0.7209 & 4.4899 & 1.718 & 0.319 & 0.7873 & 0.9117 & 0.9387  \\
    DPS & 0.1590 & 0.3341 & 0.1564 & 0.4516 & 0.1958& 0.8087 & 0.9363 & 0.9787		 \\
    DPS-FT &	0.1274 & 0.2858 & 0.0855 & 0.3815 & 0.1768 & 0.8396 & 0.9459 & 0.9866	\\
    Ours &	\textbf{0.1245} & \textbf{0.2662} & \textbf{0.0741}  & \textbf{0.3602} & \textbf{0.1666}& \textbf{0.8551} & \textbf{0.9728} & \textbf{0.9902} \\
    \hline
  \end{tabular}
  \label{sun3d}
\end{table}

We evaluate the total number of multiply-accumulate operations (MACs) needed for our approach. For a 2 image sequence, we perform 16.57 Giga Macs (GMacs) for the point detector and descriptor module, less than 0.002 GMacs for the matcher and triangulation module, and 67.90 GMacs for the sparse-to-dense module. A large fraction of this is due to the U-Net style feature tensors connecting the image and sparse depth encoder to the decoder. We perform a total of 84.48 GMacs to estimate the depth for a 2 image sequence. This is considerably lower than DPSNet which performs 295.63 GMacs for a 2 image sequence, and also less than the real-time MVDepthNet which performs 134.8 GMacs for a pair of images to estimate depth. It takes 90 milliseconds to estimate depth on Nvidia Titan RTX GPU, which we evaluated to be 2.5 times faster than DPSNet. Inference time for MVDepthNet and GPMVSNet is $\approx 60$ milliseconds. We believe our method can be further sped up by replacing Pytorch's native SVD with a custom  implementation for triangulation. Furthermore, as we do not depend on a cost volume, compound scaling laws as those derived for image \cite{tan2019efficientnet} and object \cite{tan2019efficientdet} recognition can be straightforwardly extended to our method.


\section{Conclusion}
In this work we developed an efficient depth estimation algorithm by learning to triangulate and densify sparse points in a multi-view stereo scenario. On all of the existing benchmarks, we have exceeded the state-of-the-art results, and demonstrated computation efficiency over competitive methods. In future work, we will expand on incorporating more effective attention mechanisms for interest point matching, and more anchor supporting view selection. Jointly learning depth and the full scene holistically using truncated signed distance function (TSDF) or similar representations is another promising direction. Video depth estimation approaches such as \cite{Luo-VideoDepth-2020} are closely related to MVS, and our approach can be readily extended to predict consistent and efficient depth for videos. Finally, we look forward to deeper integration with the SLAM problem, as depth estimation and SLAM are duals of each other. Overall, we believe that our approach of coupling geometry with the power of conventional 2D CNNs is a promising direction for learning 3D Vision. 

\par\vfill\par

\clearpage
%
%
\bibliographystyle{splncs04}
\bibliography{egbib}

\clearpage

\title{Supplementary Material for DELTAS: Depth Estimation by Learning Triangulation And densification of Sparse points}

\titlerunning{Depth by Triangulation and Densification}
\author{Ayan Sinha\inst{1} \and
Zak Murez\inst{1} \and
James Bartolozzi \inst{*} \and Vijay Badrinarayanan \inst{2*} \and Andrew Rabinovich \inst{3*}}
\authorrunning{A. Sinha et al.}
%
\institute{Magic Leap Inc., CA, USA \email{\{asinha,zmurez\}@magicleap.com}\\
*Work done at Magic Leap \email{bartolozzij@gmail.com} \and
Wayve.ai, London, UK \email{vijay@wayve.ai} 
\and
InsideIQ Inc., CA, USA \email{andrew@insideiq.team}}

\maketitle

\section{Ablation Studies}

We first analyze the sparse triangulated points output from our network and perform ablation studies on our critical hyper-parameters and its influence on the depth estimation performance of our trained model, i.e., we only investigate the parameter values at inference time. 

\vspace{4mm}
\noindent
\textbf{Sparse Depth Analysis:} We first validate the need to triangulate points in a differentiable manner as opposed to directly using sparse points output by a standard SLAM systems for the task of dense depth estimation. Table \ref{tablecolmap} lists the performance of sparse and dense depth estimations using COLMAP. We see that the sparse depth is of very poor quality, and the dense depth calculated by COLMAP is able to reduce the performance gap due to additional post processing steps like block matching etc. However, the best performance is that of using sparse map predicted by COLMAP as input to a sparse-to-dense depth estimation network, illustrating the power of deep networks. The sparse-to-dense network is identical to the network structure described in the main manuscript minus the triangulation module, and the descriptor and detector heads. Note that the performance of sparse-to-dense network is significantly worse than that of our approach end-to-end approach described in the main manuscript. Table \ref{tablesparse} shows the performance of the sparse depth output by the triangulation module. We see that the performance is significantly better than that of sparse points output by COLMAP, and robust accross different ratios of interest points and random points. This indicates that the network learns context around a point to circumvent the hardness of triangulating non-interest points.

\vspace{4mm}
\noindent
\textbf{Number of Points:} We first study the influence of the number of sampled points in the target image on the final depth estimation. In Table \ref{tablepoints} we see that the performance of our approach is fairly robust in the range of 256 to 512 points. Performance slightly degrades for more than 512 points. Unsurprisingly, the performance significantly degrades when no triangulated points are considered for depth estimation which would be equivalent to monocular depth estimation. However,even as few as 32 points greatly improves the performance of depth estimation. If we were to swap the depth of triangulated points with ground truth depth, we see that the performance is significantly better. Consequently, our method can be used in conjunction with an active sensor when available without requiring any retraining of the networks. A hybrid system consisting of an active sensor and our passive sensing approach is useful towards reducing the frame rate of the active sensor, and hence, reducing the power consumption. 

\vspace{4mm}
\noindent
\textbf{Ratio of points:} We investigated the influence of the ratio of the number of interest points from the interest point detector to the total number of points which are a combination of those detected by the detector and points sampled randomly from the image. For e.g., 0.75 indicates $3/4^{th}$ points sampled from the detector and the rest chosen randomly. We see in Table \ref{tableratio} that the performance of our approach is robust across all ratios. This indicates that the network is not biased towards corner points, but can robustly match points across the image. 

\vspace{4mm}
\noindent
\textbf{NMS Radius:} Next we investigate the influence of the non-maximum suppression (NMS) radius value for the interest point detector on the performance. Note that small values of NMS result in interest points being sampled predominantly from high texture regions and being clustered together, whereas high values of NMS encourage the points to be well distributed. In Table \ref{tablenms} we see that small values of NMS hurt performance, with the performance improving till NMS value of 9 and then again degrading for value of 11. This indicates that the network prefers well separated, uniformly sampled points across the image.   

\vspace{4mm}
\noindent
\textbf{Threshold:} We also investigated the performance of depth estimation for different thresholds on the interest point detector. In Table \ref{tablethresh} we see that threshold values of 0.0001 and 0.0005 result in similar performance. The performance degrades for higher values of 0.001 and 0.005. This suggests that the network does not particularly favour high quality interest points, but a large number of them, which are made available when the threshold is low. 

\vspace{4mm}
\noindent
\textbf{Epipolar Length:} In Table \ref{tablelength} we investigate the influence of the length of the sampled descriptors along the epipolar line on depth estimation. We see that the performance is robust across all values of length ranging from 25 pixels to 150 pixels. This observation can further reduce the training time and inference time for depth estimation. 

\begin{table}[ht]
  \caption{Performance of depth estimation on ScanNet using COLMAP. Sparse refers to the sparse map predicted by COLMAP, Dense refers to the dense depth map predicted by COLMAP, and Sparse +DNN refers to densification of the sparse map predicted by COLMAP using a deep neural network. 
  }
  \centering
  \begin{tabular}{ccccccccc}
 \hline 
   Approach  &Abs Rel  &Abs & Sq Rel & \textbf{$\delta< 1.25$}& \textbf{$\delta < 1.25^2$}& \textbf{$\delta< 1.25^3$}\\
    \hline
Sparse &	0.2629 &	0.4618 &	0.3882 &	0.5713 &	0.7498 &	0.8322\\
\hline
Dense &	0.1371 & 0.2643 & 0.1379 & 0.8344 & 0.9080 & 0.9383  \\
Sparse + DNN  &	\textbf{0.1242} & \textbf{0.1990} & \textbf{0.0658}  &  \textbf{0.8756} & \textbf{0.9649} & \textbf{0.9878} \\

\hline
  \end{tabular}
\label{tablecolmap}
\end{table}

\begin{table}[ht]
  \caption{Performance of sparse depth estimation on ScanNet for different ratios of interest points and random points. We use sequences of length 3 and sample every 20 frames.
  }
  \centering
  \begin{tabular}{ccccccccc}
 \hline 
   Ratio  &Abs Rel  &Abs & Sq Rel & \textbf{$\delta< 1.25$}& \textbf{$\delta < 1.25^2$}& \textbf{$\delta< 1.25^3$}\\
    \hline
0.0 &	0.0993 &	0.1899 &	0.0503 &	0.8856 &	0.9701 &	\textbf{0.9906}\\
0.25 & \textbf{0.0986} &	\textbf{0.1891}&	\textbf{0.0502} &	\textbf{0.8869} &	\textbf{0.9703} &	\textbf{0.9906}\\
0.5 &	0.0988 & 0.1893 & 0.0503 &  0.8866 & 0.9702 & 0.9905 \\
0.75 &	0.0988 & 0.1893 & 0.0503  &  0.8866 & 0.9702 & 0.9905 \\
1.0 &	0.0988 & 0.1893 &0.0503  &  0.8866 & 0.9702 & 0.9905 \\
\hline
  \end{tabular}
\label{tablesparse}
\end{table}

\begin{table}[ht]
  \caption{Performance of depth estimation on ScanNet for different number of sparse points. We use sequences of length 3 and sample every 20 frames. }
  \centering
  \begin{tabular}{ccccccccc}
 \hline 
   Num Points  &Abs Rel  &Abs & Sq Rel & \textbf{$\delta< 1.25$}& \textbf{$\delta < 1.25^2$}& \textbf{$\delta< 1.25^3$}\\

    \hline
0 &	0.2203 &	0.3049 &	0.1375 &	0.7198 &	0.9022 &	0.9650\\
32 &	0.1105 &	0.1793 &	0.0582 &	0.9002 &	0.9722 &	0.9886\\
128 &	0.0960 &	0.1591 &	0.0514 &	0.9232&	0.9760&	\textbf{0.9895}\\
256 &	0.0934&	0.1550&	\textbf{0.0505}&	0.9276&	0.9766&	\textbf{0.9895}\\
384 &	\textbf{0.0931}&	0.1541&	\textbf{0.0505}&	0.9285&	\textbf{0.9767}&	0.9894\\
512 &	0.0932 & \textbf{0.1540} & 0.0506 &  \textbf{0.9287} & \textbf{0.9767} & 0.9893 \\
640 &	0.0936&	0.1543&	0.0509&	0.9285&	0.9766&	0.9892\\
768 &	0.0942&	0.1549&	0.0512&	0.9282&	0.9766&	0.9891\\
\hline
512 (GT) &	\textbf{0.0680} & \textbf{0.1111} & \textbf{0.0406} &  \textbf{0.9562} & \textbf{0.9800} & \textbf{0.9903} \\
\hline
  \end{tabular}
\label{tablepoints}
\end{table}


\begin{table}[h]
  \caption{Performance of depth estimation on ScanNet for different ratios of interest points and random points. We use sequences of length 3 and sample every 20 frames. }
  \centering
  \begin{tabular}{ccccccccc}
 \hline 
   Ratio  &Abs Rel  &Abs & Sq Rel & \textbf{$\delta< 1.25$}& \textbf{$\delta < 1.25^2$}& \textbf{$\delta< 1.25^3$}\\
    \hline
0.0 &	0.0935&	0.1544&	0.0508&	0.9283&	0.9766&	\textbf{0.9893}\\
0.25 &0.0933&	\textbf{0.1540}&	0.0507&	0.9286&	0.9766&	\textbf{0.9893}\\
0.5 &	\textbf{0.0932} & \textbf{0.1540} & \textbf{0.0506} &  \textbf{0.9287} & \textbf{0.9767} & \textbf{0.9893} \\
0.75 &	0.0933& \textbf{0.1540}& 0.0507& 0.9286& 0.9766& \textbf{0.9893}\\
1.0 &	0.0933&	\textbf{0.1540}&	0.0507&	0.9286&	0.9766&	\textbf{0.9893}\\
\hline
  \end{tabular}
\label{tableratio}
\end{table}


\begin{table}[h]
  \caption{Performance of depth estimation on ScanNet for different radius for non-maximum suppression (NMS Rad). We use sequences of length 3 and sample every 20 frames. }
  \centering
  \begin{tabular}{ccccccccc}
 \hline 
   NMS Rad  &Abs Rel  &Abs & Sq Rel & \textbf{$\delta< 1.25$}& \textbf{$\delta < 1.25^2$}& \textbf{$\delta< 1.25^3$}\\
    \hline
3 &	0.0942&	0.1554&	0.0512&	0.9278&	0.9765&	0.9892\\
5 & 0.0937&	0.1545&	0.0508&	0.9283&	0.9766&	0.9892\\
7 &0.0937&	0.1545&	0.0510&	0.9284&	0.9766&	0.9892\\
9 &	\textbf{0.0932} & \textbf{0.1540} & \textbf{0.0506} &  \textbf{0.9287} & \textbf{0.9767} & \textbf{0.9893} \\
11 &0.0938& 0.1546&	0.0511&	0.9285&	0.9766&	0.9891\\
\hline
  \end{tabular}
\label{tablenms}
\end{table}

 
\begin{table}[h]
  \caption{Performance of depth estimation on ScanNet for different thresholds for the detector. We use sequences of length 3 and sample every 20 frames. }
  \centering
  \begin{tabular}{ccccccccc}
 \hline 
   Thresh  &Abs Rel  &Abs & Sq Rel & \textbf{$\delta< 1.25$}& \textbf{$\delta < 1.25^2$}& \textbf{$\delta< 1.25^3$}\\
    \hline
0.0001 & \textbf{0.0932} & \textbf{0.1539} & \textbf{0.0506} &  \textbf{0.9287} & \textbf{0.9767} & \textbf{0.9893} \\
0.0005 & \textbf{0.0932} & 0.1540 & \textbf{0.0506} &  \textbf{0.9287} & \textbf{0.9767} & \textbf{0.9893} \\
0.001 &0.0933&	0.1540&	0.0507&	0.9286&	\textbf{0.9767}&	\textbf{0.9893}\\
0.005 & 0.0934&	0.1544&	0.0507&	0.9283&	0.9766&	\textbf{0.9893}\\

\hline
  \end{tabular}
\label{tablethresh}
\end{table}


\begin{table}[h]
  \caption{Performance of depth estimation on ScanNet for different lengths of the sampled descriptors. We use sequences of length 3 and sample every 20 frames. }
  \centering
  \begin{tabular}{ccccccccc}
 \hline 
   Length  &Abs Rel  &Abs & Sq Rel & \textbf{$\delta< 1.25$}& \textbf{$\delta < 1.25^2$}& \textbf{$\delta< 1.25^3$}\\
    \hline
25 & 0.0934&	0.1542&	0.0508&	\textbf{0.9287}&	\textbf{0.9767}&	\textbf{0.9893}\\
50 & 0.0933&\textbf{0.1540}&	0.0507&	\textbf{0.9287}&	\textbf{0.9767}&	\textbf{0.9893}\\
100 & \textbf{0.0932} & \textbf{0.1540} & \textbf{0.0506} &  \textbf{0.9287} & \textbf{0.9767} & \textbf{0.9893} \\
150 &\textbf{0.0932}&	\textbf{0.1540}&	\textbf{0.0506}&	0.9286&	\textbf{0.9767}&	\textbf{0.9893}\\

\hline
  \end{tabular}
\label{tablelength}
\end{table}


\begin{table}[t]
  \caption{Performance of depth estimation on ScanNet for different sampling offsets. We use sequences of length 3 and sample every 20 frames. }
  \centering
  \begin{tabular}{ccccccccc}
 \hline 
   Offsets  &Abs Rel  &Abs & Sq Rel & \textbf{$\delta< 1.25$}& \textbf{$\delta < 1.25^2$}& \textbf{$\delta< 1.25^3$}\\

    \hline

1 pix &	\textbf{0.0932} & \textbf{0.1540} & \textbf{0.0506} &  \textbf{0.9287} & \textbf{0.9767} & \textbf{0.9893} \\
2 pix &0.0933&	0.1541&	0.0507&	0.9285&	0.9766&	\textbf{0.9893}\\

\hline
  \end{tabular}
\label{tableoff}
\end{table}

\begin{table}[t]
  \caption{Performance of depth estimation on ScanNet for different architectures. We use sequences of length 3 and sample every 20 frames. }
  \centering
  \begin{tabular}{cccccccccc}
 \hline 
   Arch  &Abs Rel  &Abs & Sq Rel & \textbf{$\delta< 1.25$}& \textbf{$\delta < 1.25^2$}& \textbf{$\delta< 1.25^3$} & GMACs\\
    \hline
MVDepth & 0.1054 & 0.1911 & 0.0970 & 0.8952 & 0.9707 & 0.9895 & 134.8  \\
DPS &	0.1025 & 0.1675 & 0.0574 &  0.9102 & 0.9708 & 0.9872 & 295.6	\\
VGG-9&0.1073&	0.1815&	0.0581&	0.9023&	0.9719&	0.9890 & \textbf{16.9} \\
ResNet-50 &	\textbf{0.0932} & \textbf{0.1540} & \textbf{0.0506} &  \textbf{0.9287} & \textbf{0.9767} & \textbf{0.9893} & 84.4 \\
\hline
  \end{tabular}
\label{tablearch}
\end{table}

\begin{table}
  \caption{Performance of different descriptors on ScanNet.}
  \centering
  \begin{tabular}{ccccccccc}
  \hline
     &\textbf{MLE}  &\textbf{MScore} & \textbf{Num} & \textbf{Rep}& \textbf{rot@$5^{\circ}$}& \textbf{trans@5cm}\\

    \hline
SuperPoint  &  \textbf{2.545} &     \textbf{0.375}  &   129  & 0.519	&  0.489 &      0.244 \\
VGG-9 &	3.057 &		0.325  &   1619  & \textbf{0.751} &  0.472 &		0.228	\\
ResNet-50 &	 3.101 &   0.329   &  1511  & 0.738 &  \textbf{0.518} &\textbf{0.254} 	\\
\hline
  \end{tabular}
  \label{tabledd}
\end{table}

\vspace{4mm}
\noindent
\textbf{Offset value:} We investigated the performance of our trained network for 1 pixel and 2 pixel offsets to compensate for pose error. We see in Table \ref{tableoff} that 2 pixel offset does not improve performance, suggesting that the pose in ScanNet is sufficiently reliable. This parameter however might be of greater influence in cases wherein pose estimation is unreliable.

\vspace{4mm}
\noindent
\textbf{Model Architecture:} Finally, we explore the performance of our approach on model architecture. We swap our ResNet-50 backbone with a VGG-9 backbone similar to that of SuperPoint. We use the same training procedure as that of the ResNet-50 architecture mentioned in the main manuscript. In Table \ref{tablearch} we that the extremely light-weight VGG-9 architecture performs much better than MVDepthNet and some values are comparable or even better than those of DPSNet. Furthermore, the total number of GMACs is only 16.9, which is $\approx 18$x more efficient that DPSNet and  $8$x more efficient that real-time MVDepthNet. In Table \ref{tabledd} we see that we observe only a slight degradation in pose performance (rotation and translation) compared to SuperPoint. This reinforces our conclusion in the main manuscript that our supervision can complement that of SuperPoint. Overall, the robust performance of our network with extremely low compute is a promising first step to derive scaling laws as done in EfficientNet.    

\begin{figure}
  \centering
    \begin{tabular}{cccccc}
        \includegraphics[width=.16\textwidth,height=1.5cm]{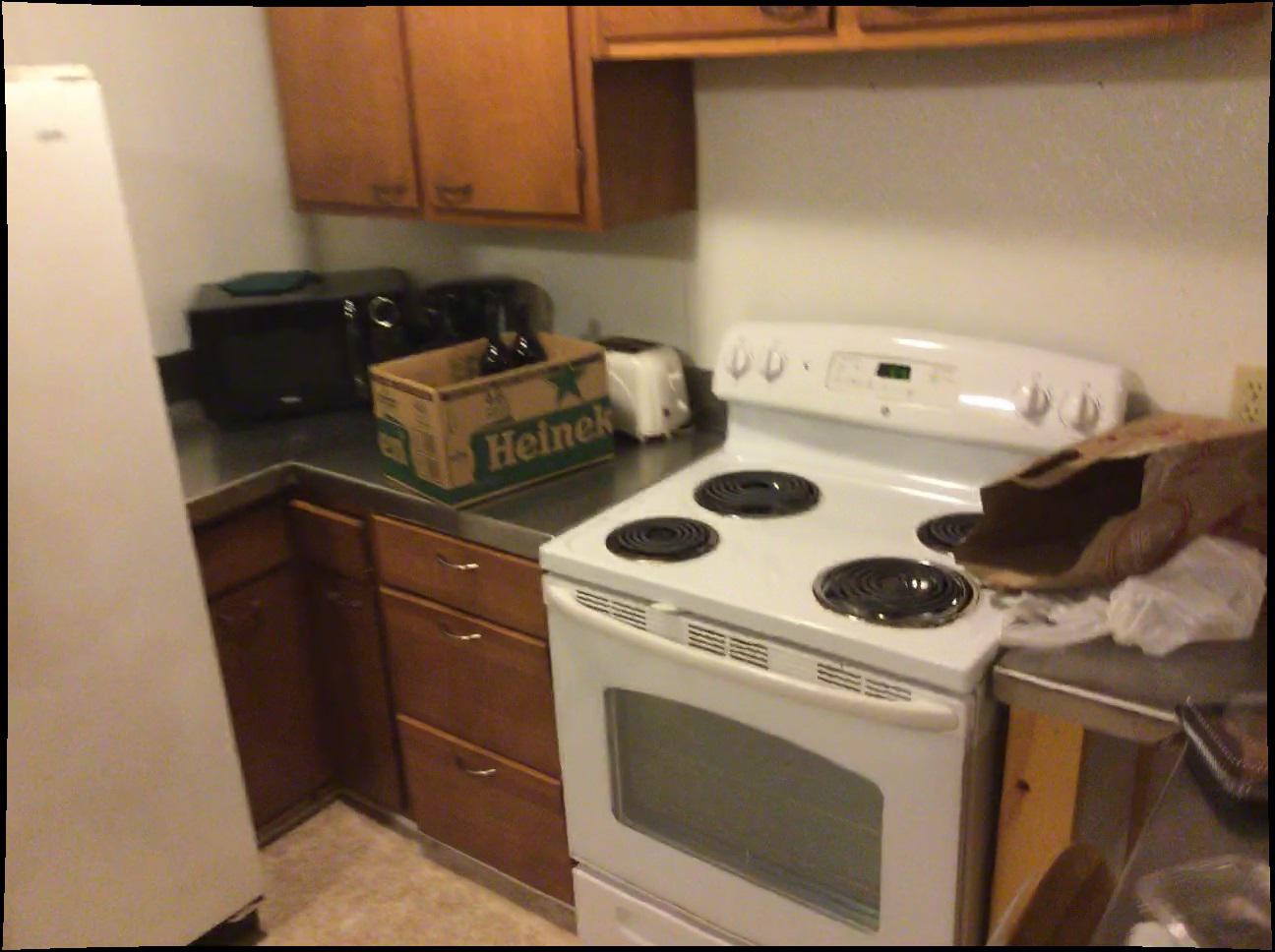} &
        \includegraphics[width=.16\textwidth,height=1.5cm]{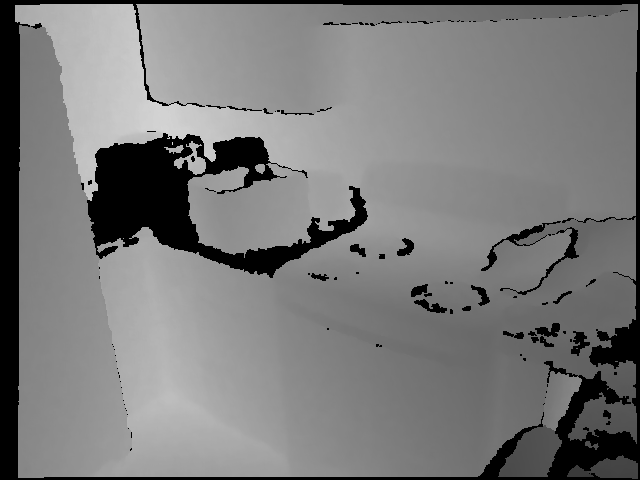} &    \includegraphics[width=.16\textwidth,height=1.5cm]{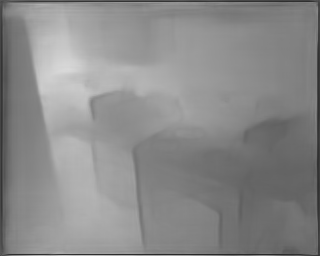} &    \includegraphics[width=.16\textwidth,height=1.5cm]{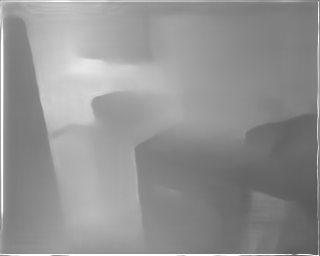} &    \includegraphics[width=.16\textwidth,height=1.5cm]{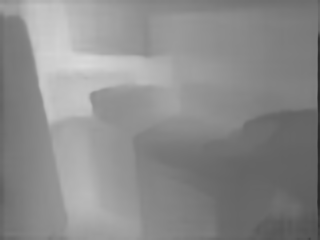} &    \includegraphics[width=.16\textwidth,height=1.5cm]{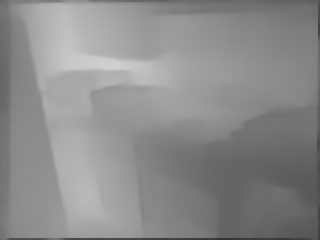} \\
        \includegraphics[width=.16\textwidth,height=1.5cm]{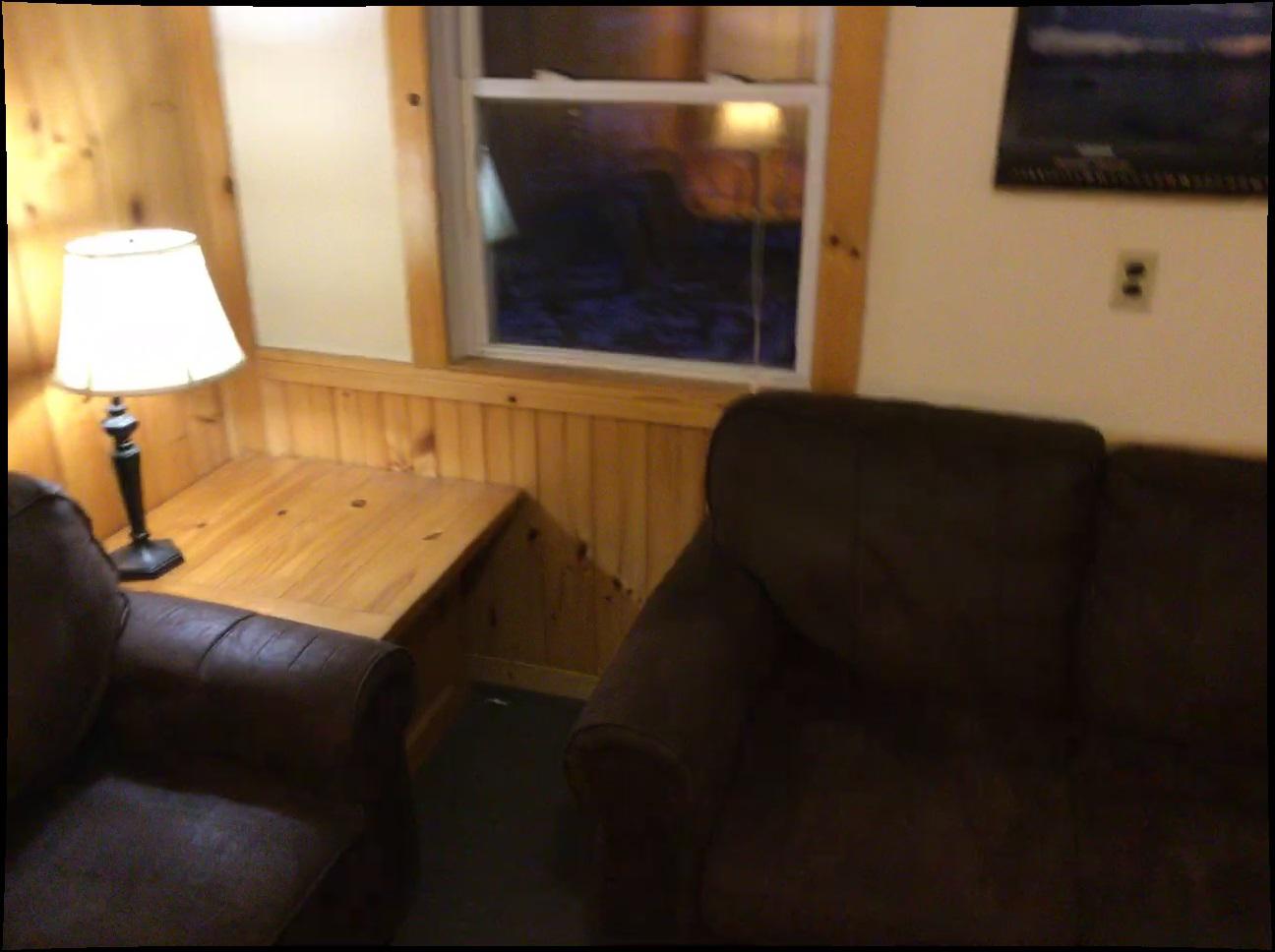} &
        \includegraphics[width=.16\textwidth,height=1.5cm]{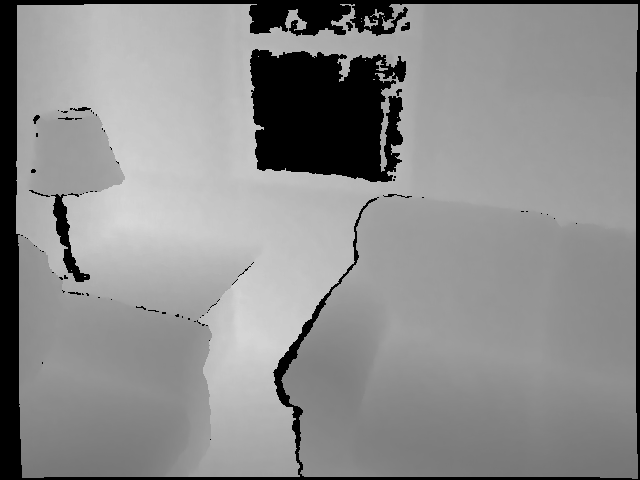} &    \includegraphics[width=.16\textwidth,height=1.5cm]{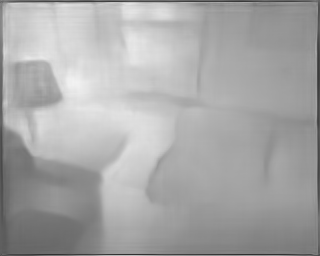} &    \includegraphics[width=.16\textwidth,height=1.5cm]{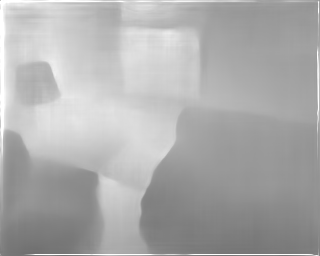} &    \includegraphics[width=.16\textwidth,height=1.5cm]{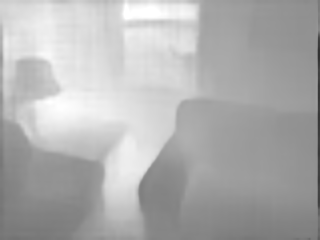} &    \includegraphics[width=.16\textwidth,height=1.5cm]{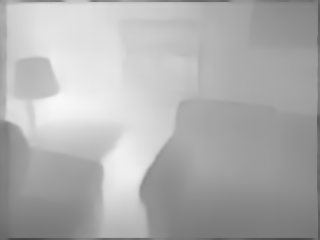} \\         \includegraphics[width=.16\textwidth,height=1.5cm]{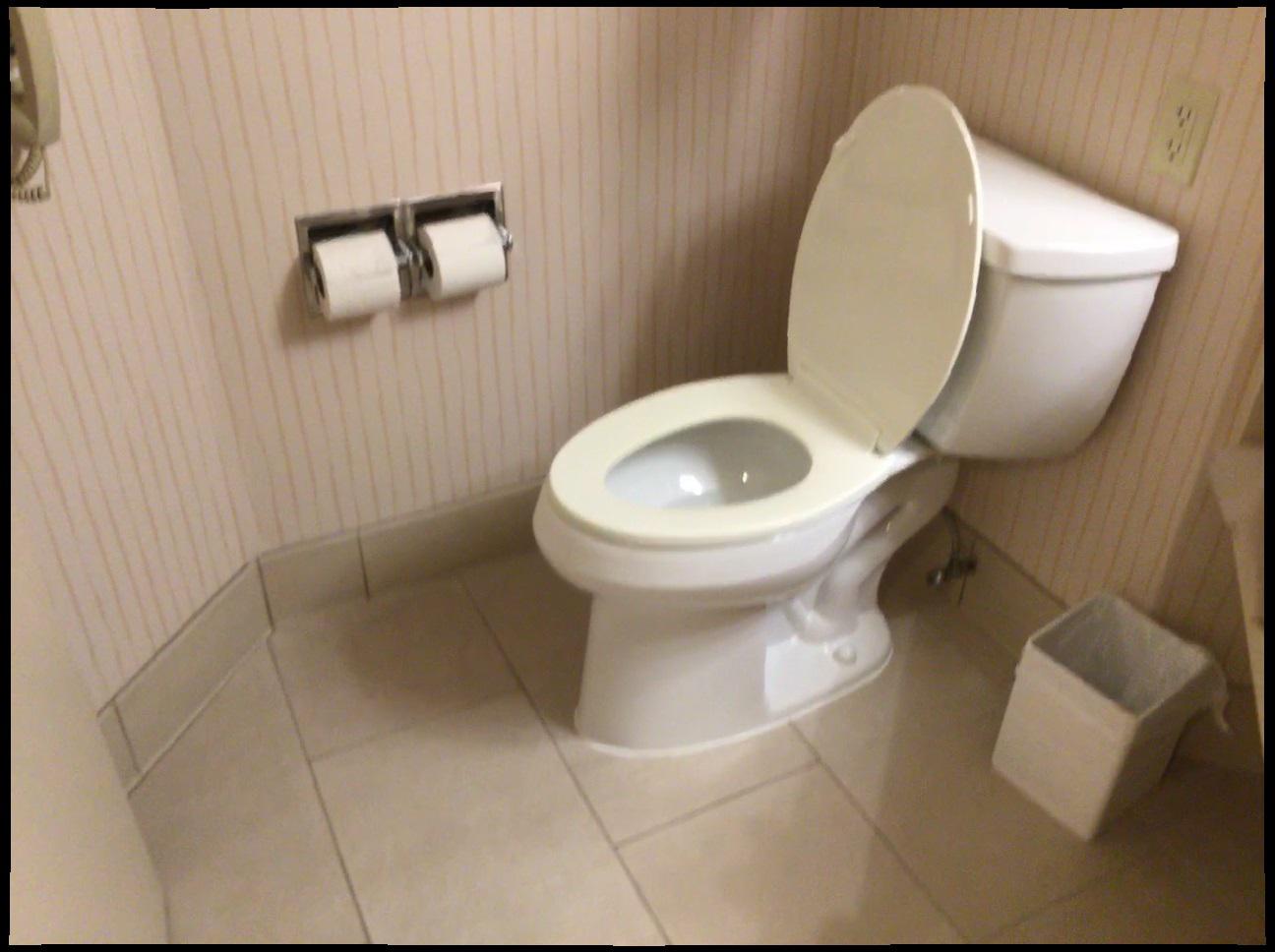} &
        \includegraphics[width=.16\textwidth,height=1.5cm]{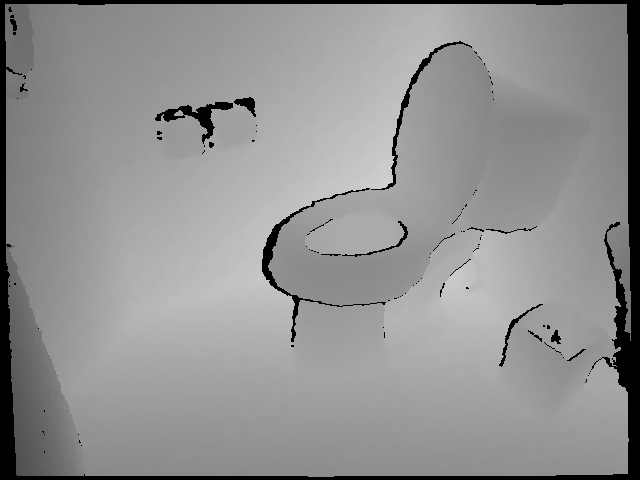} &    \includegraphics[width=.16\textwidth,height=1.5cm]{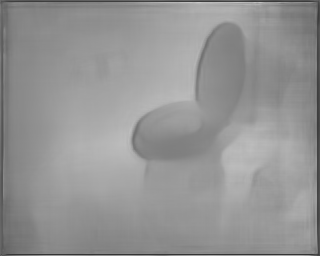} &    \includegraphics[width=.16\textwidth,height=1.5cm]{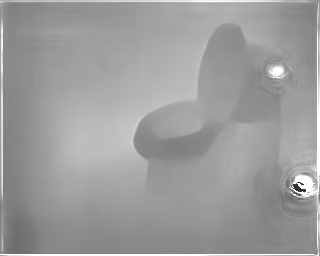} &    \includegraphics[width=.16\textwidth,height=1.5cm]{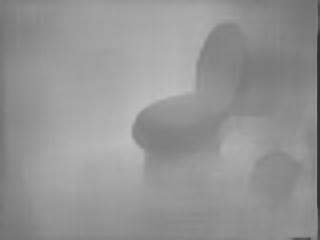} &    \includegraphics[width=.16\textwidth,height=1.5cm]{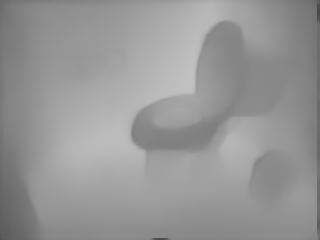} \\         \includegraphics[width=.16\textwidth,height=1.5cm]{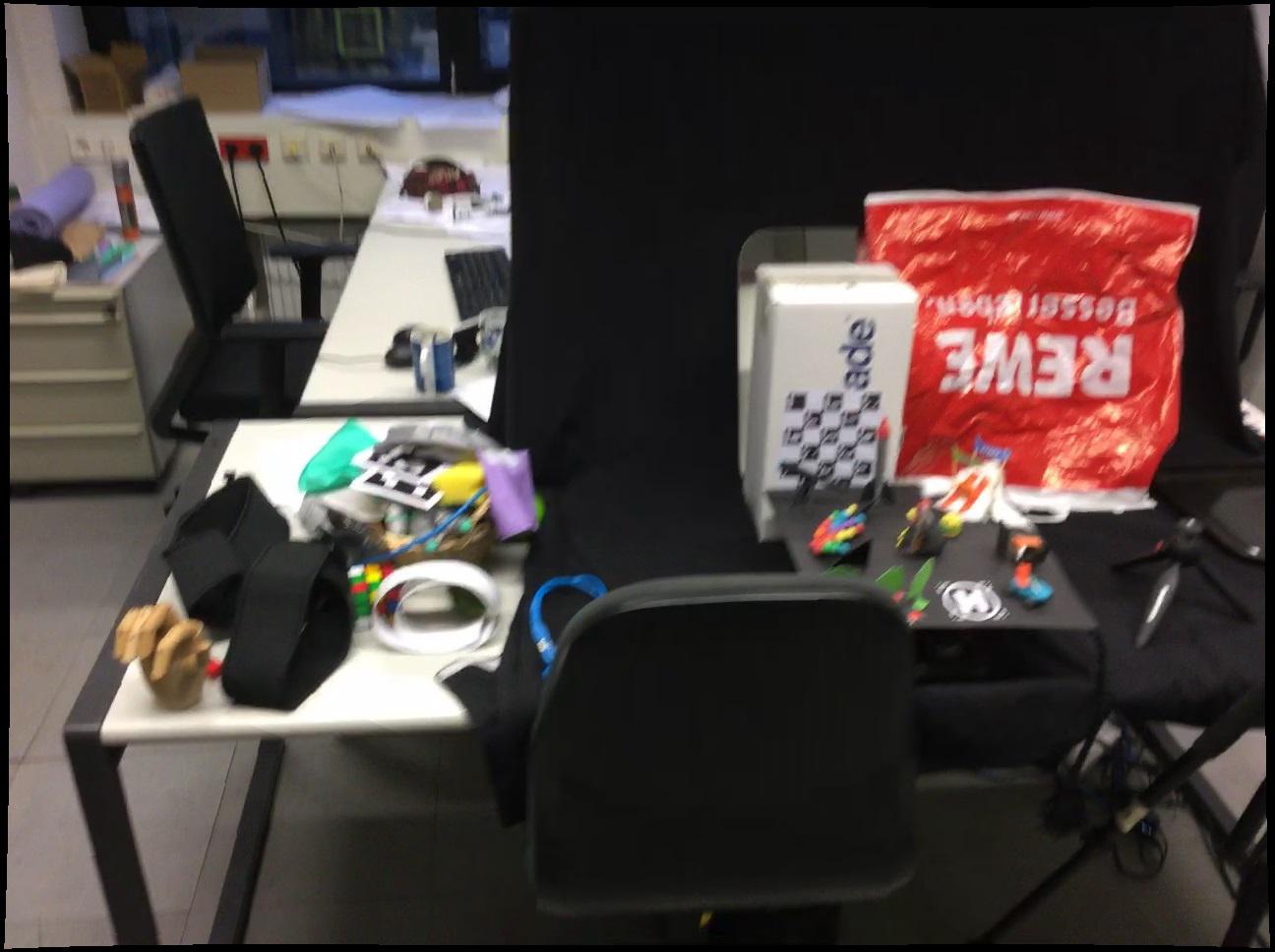} &
        \includegraphics[width=.16\textwidth,height=1.5cm]{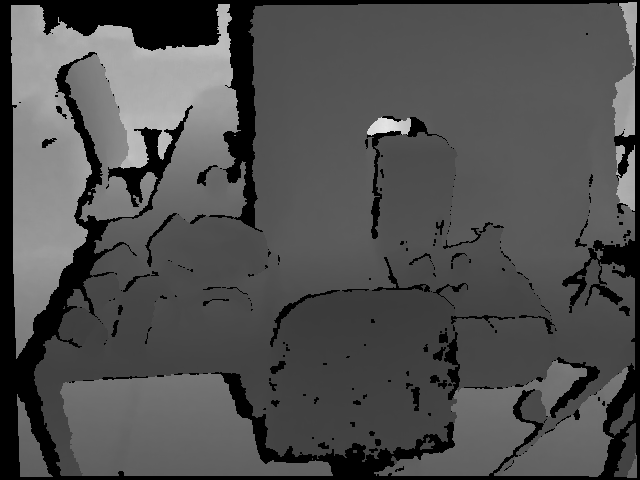} &  
        \includegraphics[width=.16\textwidth,height=1.5cm]{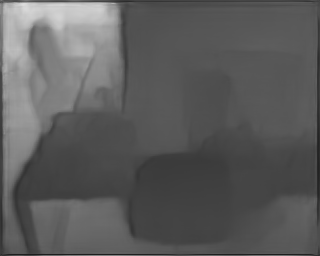} &    \includegraphics[width=.16\textwidth,height=1.5cm]{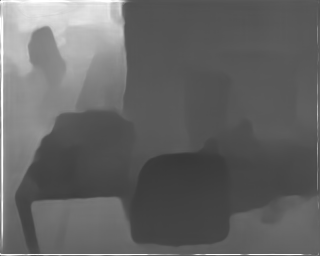} &    \includegraphics[width=.16\textwidth,height=1.5cm]{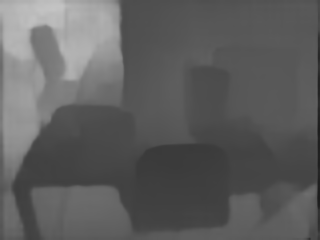} &    \includegraphics[width=.16\textwidth,height=1.5cm]{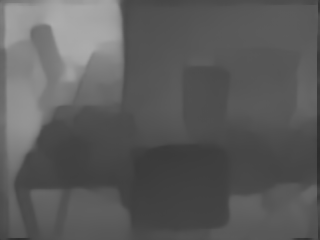} \\         \includegraphics[width=.16\textwidth,height=1.5cm]{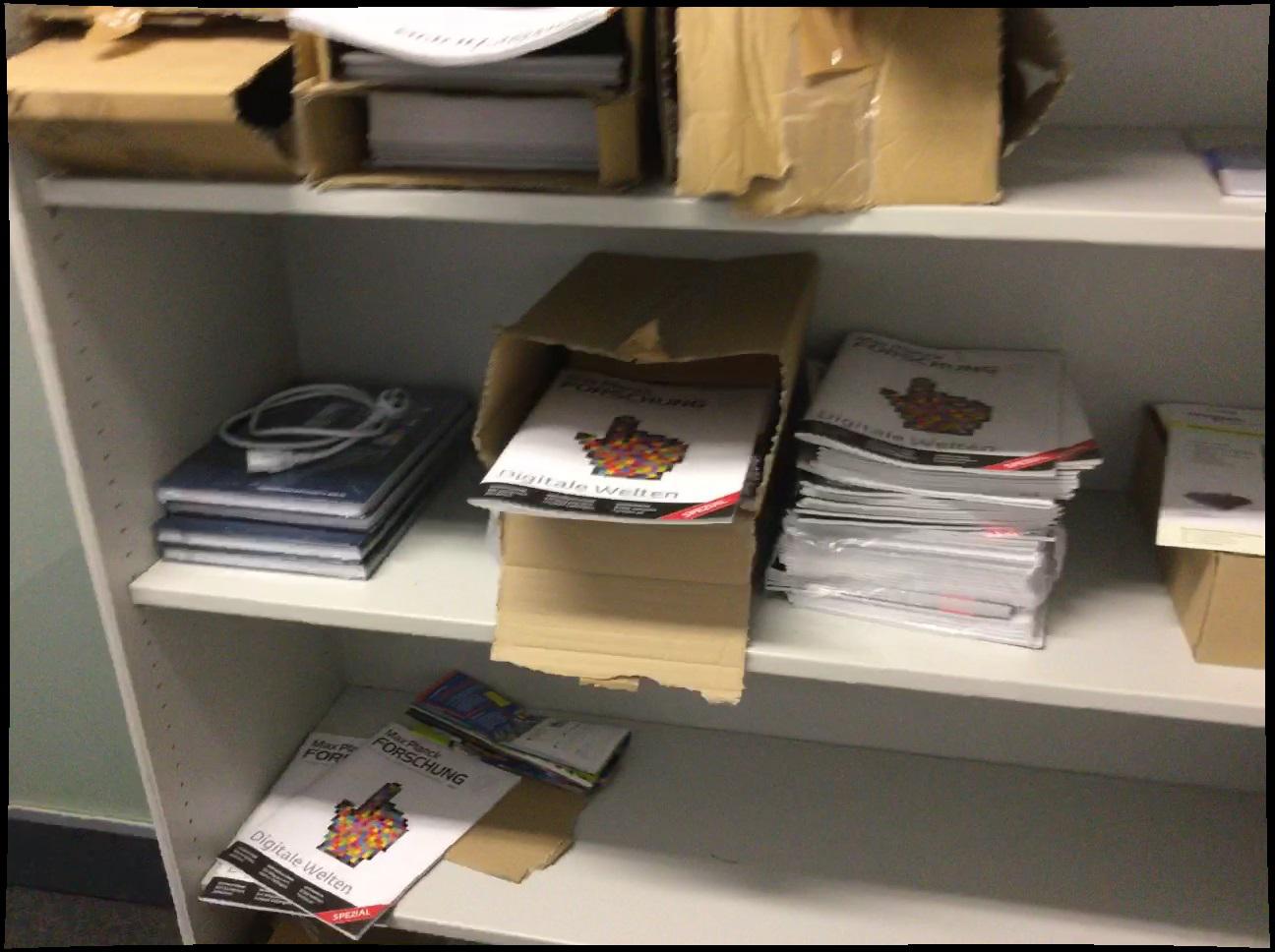} &
        \includegraphics[width=.16\textwidth,height=1.5cm]{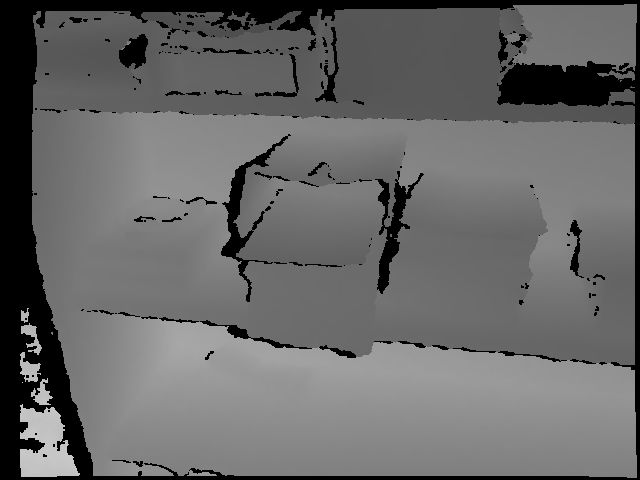} &    \includegraphics[width=.16\textwidth,height=1.5cm]{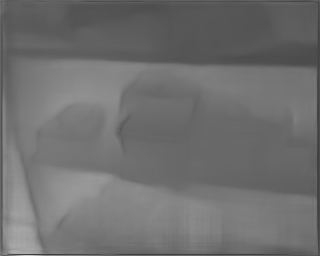} &    \includegraphics[width=.16\textwidth,height=1.5cm]{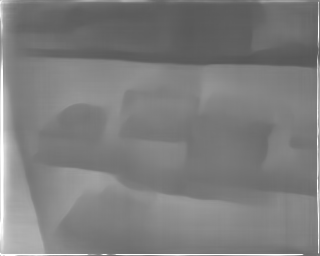} &    \includegraphics[width=.16\textwidth,height=1.5cm]{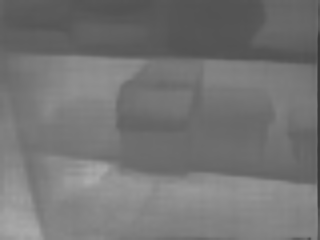} &    \includegraphics[width=.16\textwidth,height=1.5cm]{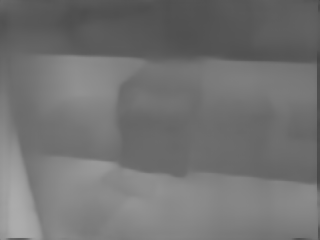} \\         \includegraphics[width=.16\textwidth,height=1.5cm]{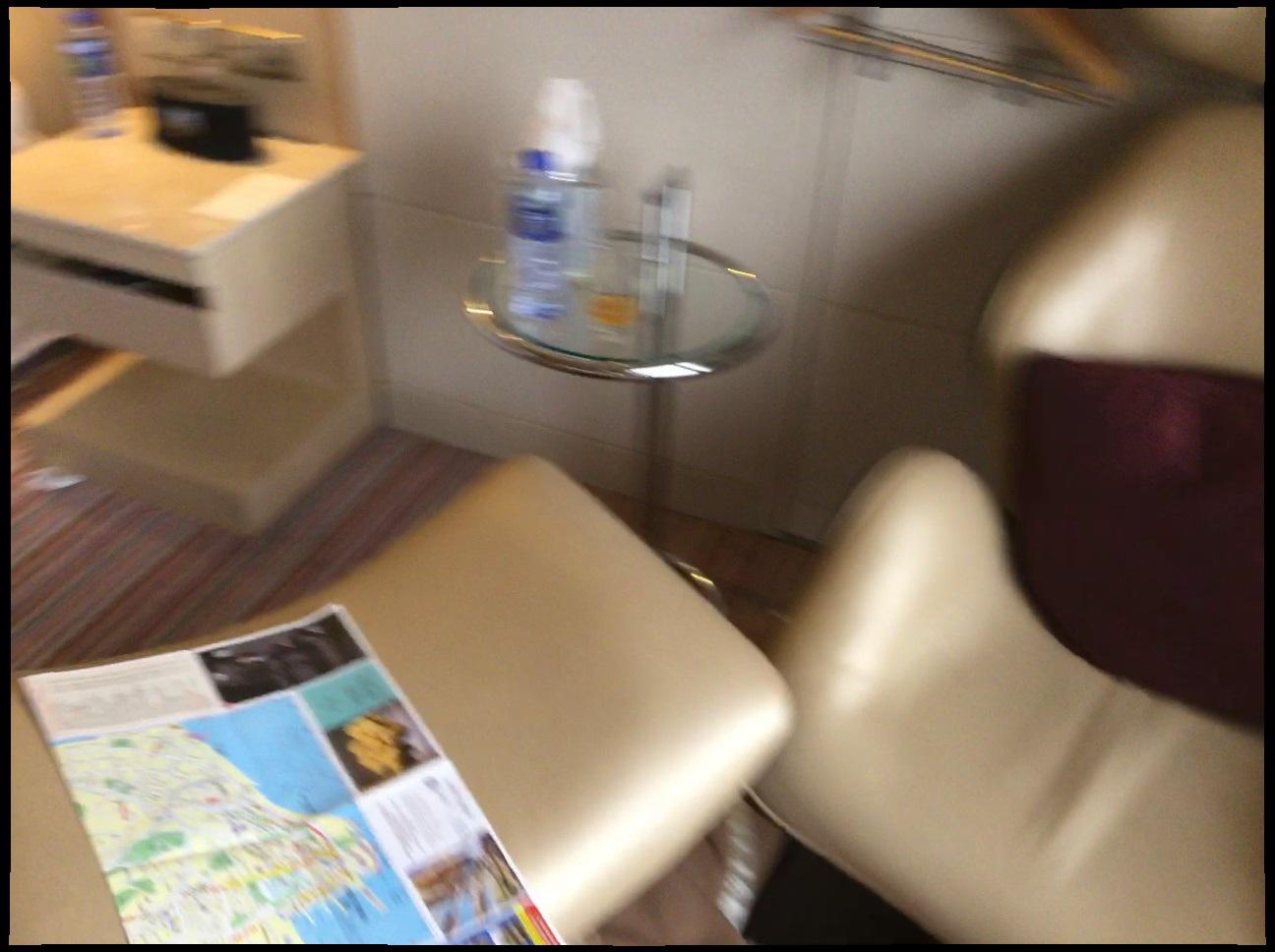} &
        \includegraphics[width=.16\textwidth,height=1.5cm]{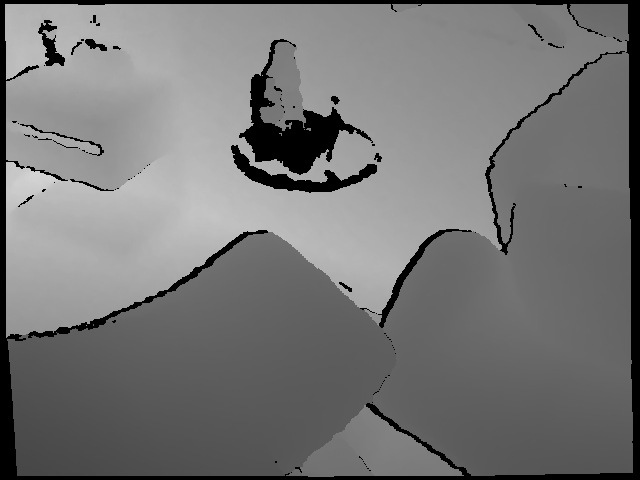} &    \includegraphics[width=.16\textwidth,height=1.5cm]{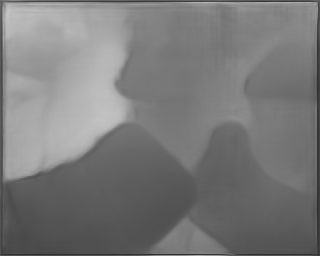} &    \includegraphics[width=.16\textwidth,height=1.5cm]{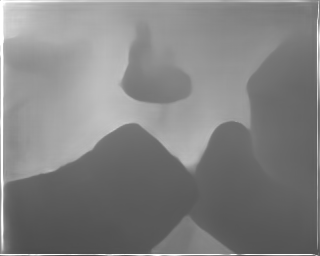} &    \includegraphics[width=.16\textwidth,height=1.5cm]{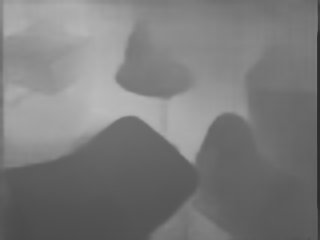} &    \includegraphics[width=.16\textwidth,height=1.5cm]{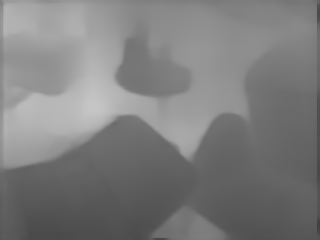}\\         \includegraphics[width=.16\textwidth,height=1.5cm]{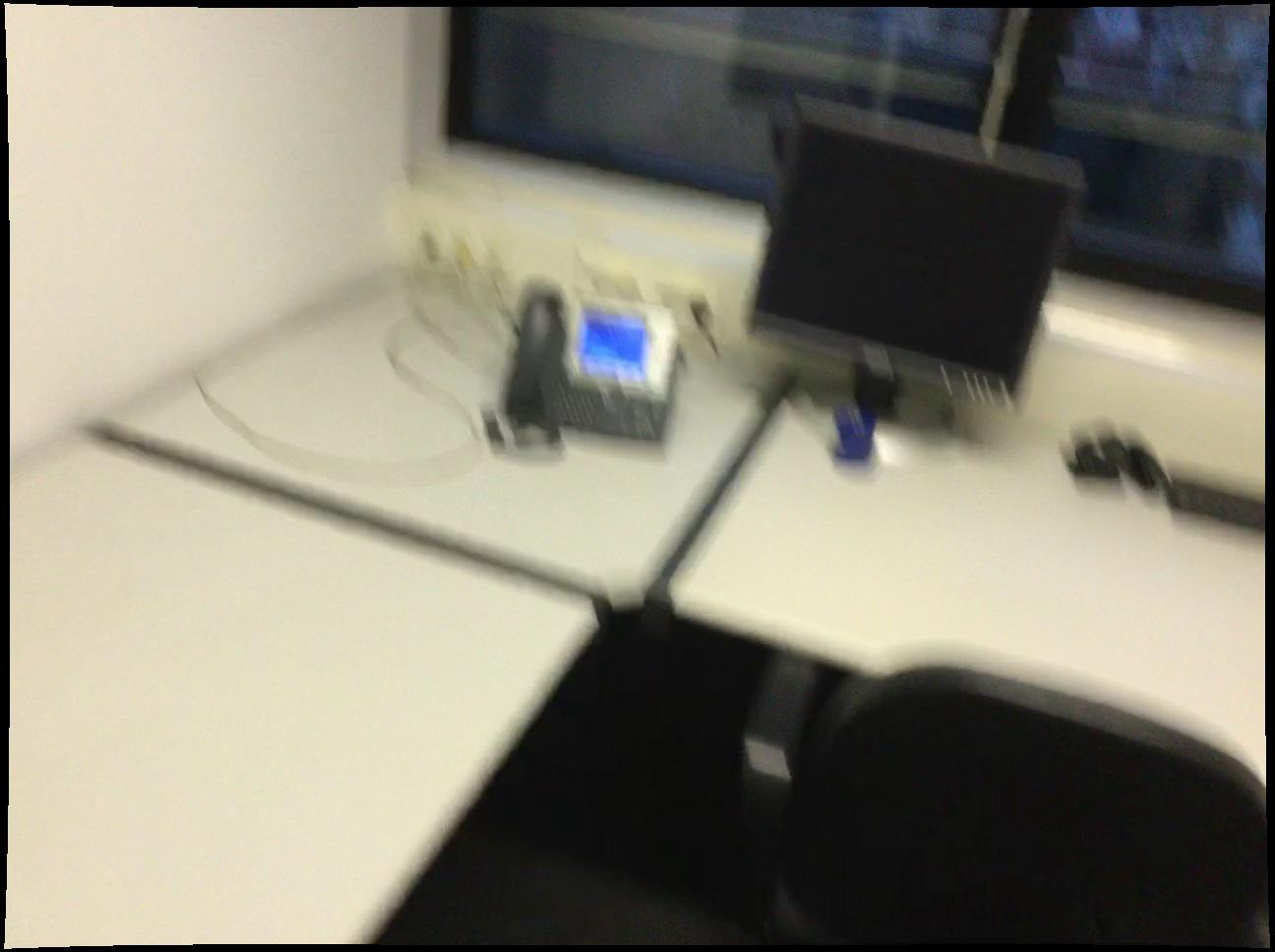} &
        \includegraphics[width=.16\textwidth,height=1.5cm]{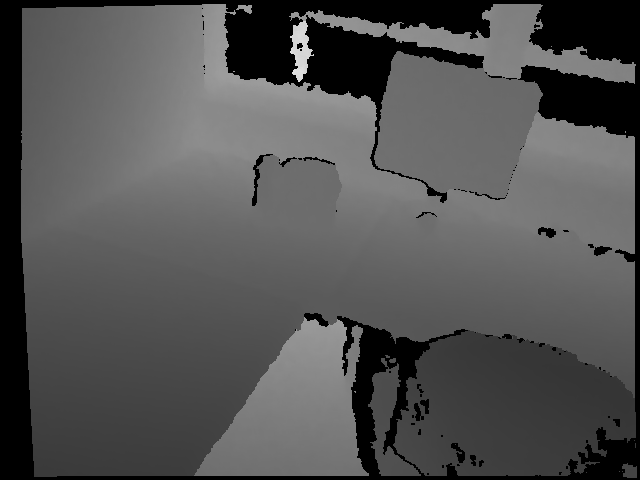} &    \includegraphics[width=.16\textwidth,height=1.5cm]{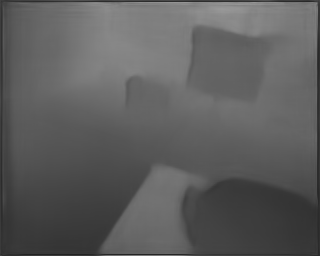} &    \includegraphics[width=.16\textwidth,height=1.5cm]{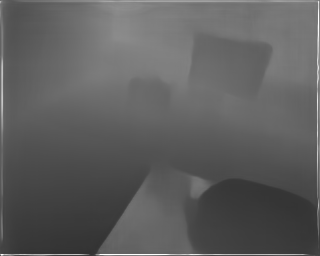} &    \includegraphics[width=.16\textwidth,height=1.5cm]{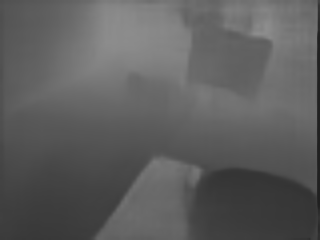} &    \includegraphics[width=.16\textwidth,height=1.5cm]{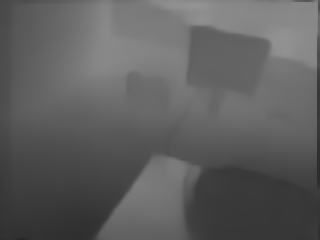} \\         \includegraphics[width=.16\textwidth,height=1.5cm]{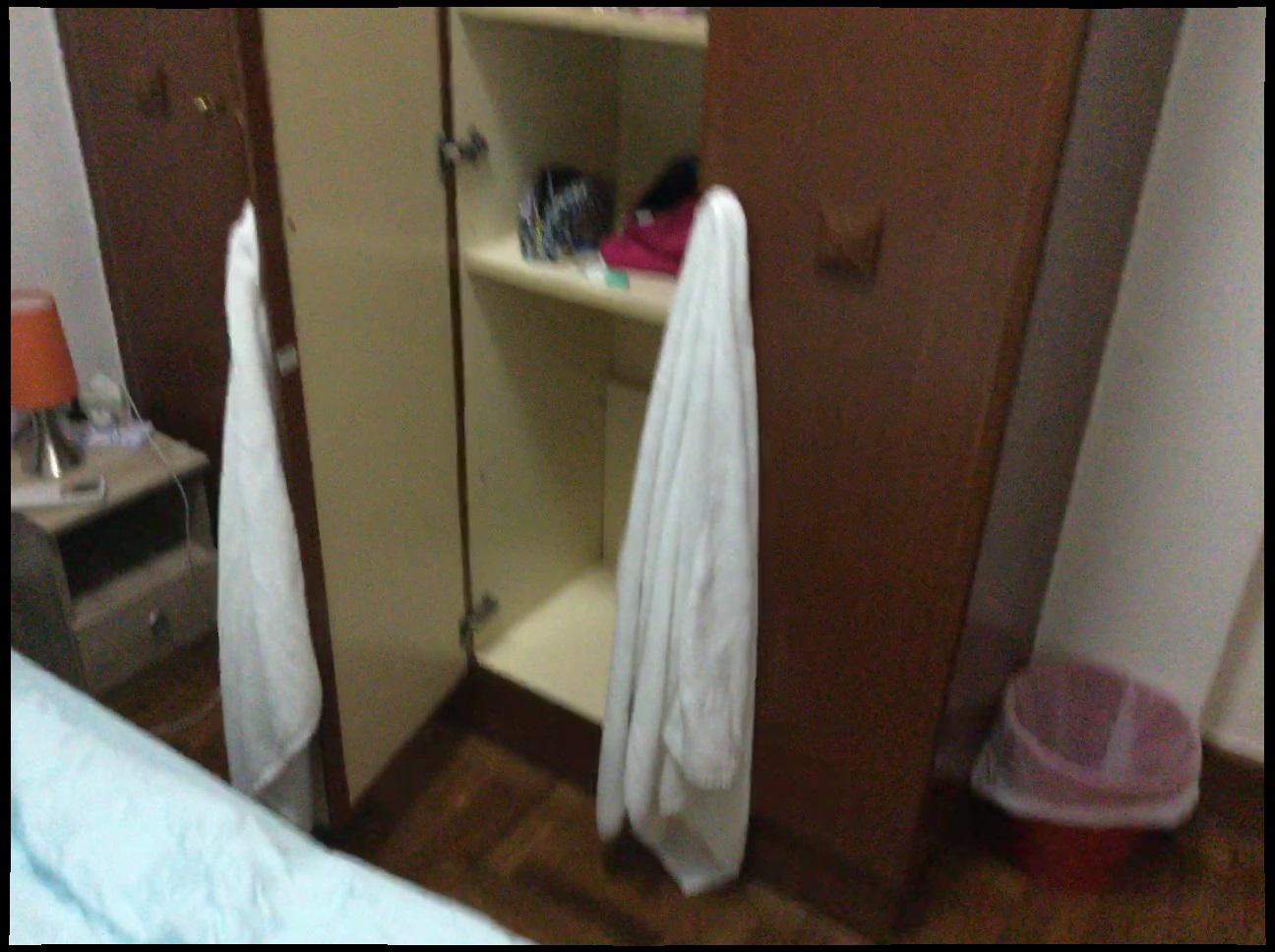} &
        \includegraphics[width=.16\textwidth,height=1.5cm]{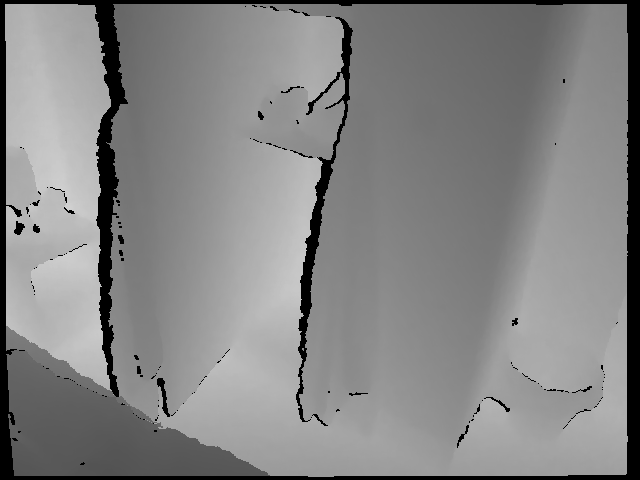} &    \includegraphics[width=.16\textwidth,height=1.5cm]{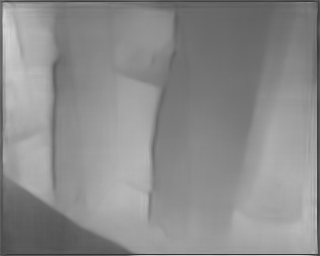} &    \includegraphics[width=.16\textwidth,height=1.5cm]{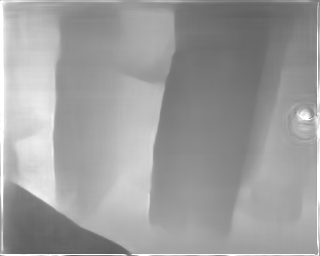} &    \includegraphics[width=.16\textwidth,height=1.5cm]{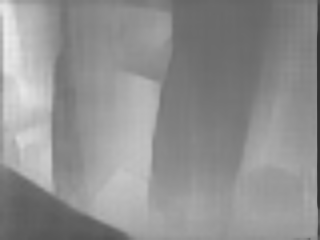} &    \includegraphics[width=.16\textwidth,height=1.5cm]{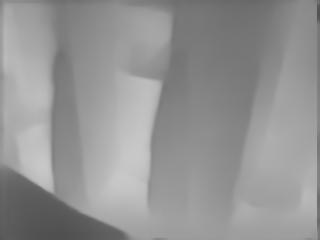} \\         \includegraphics[width=.16\textwidth,height=1.5cm]{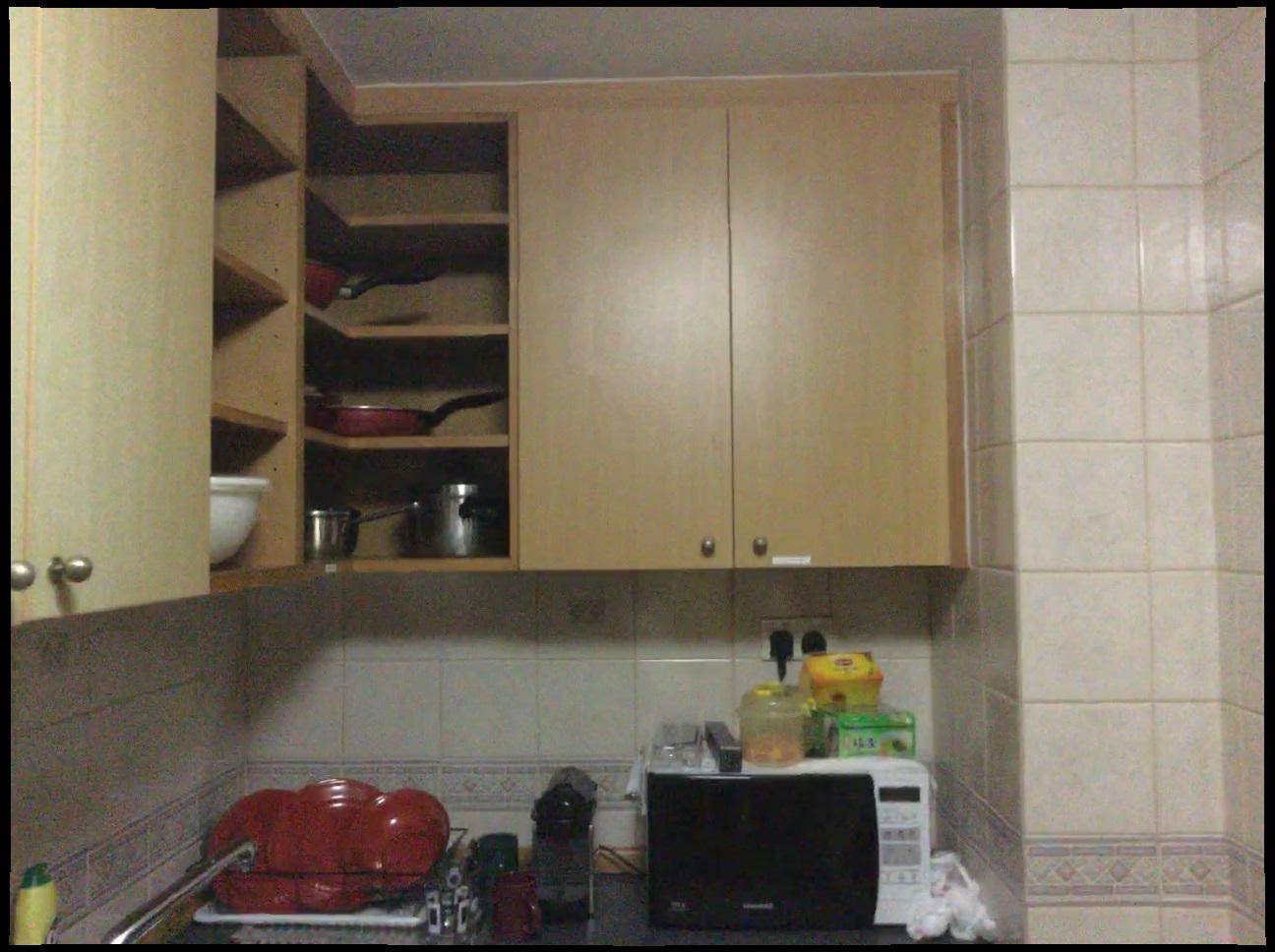} &
        \includegraphics[width=.16\textwidth,height=1.5cm]{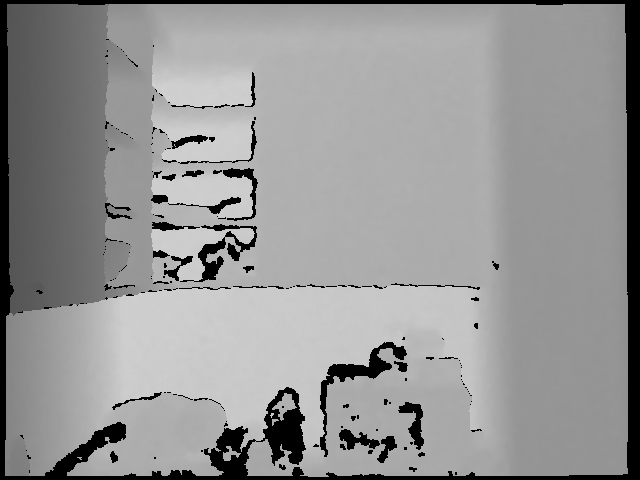} &    \includegraphics[width=.16\textwidth,height=1.5cm]{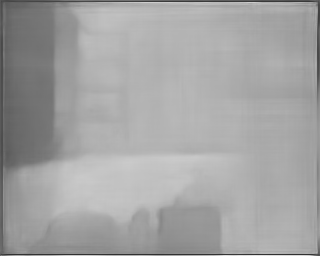} &    \includegraphics[width=.16\textwidth,height=1.5cm]{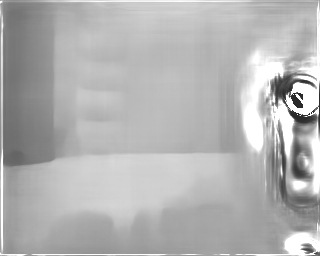} &    \includegraphics[width=.16\textwidth,height=1.5cm]{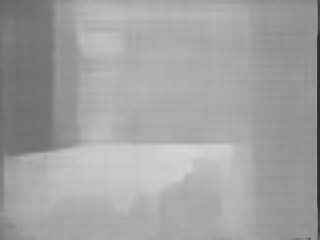} &    \includegraphics[width=.16\textwidth,height=1.5cm]{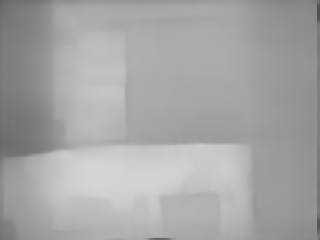} \\         \includegraphics[width=.16\textwidth,height=1.5cm]{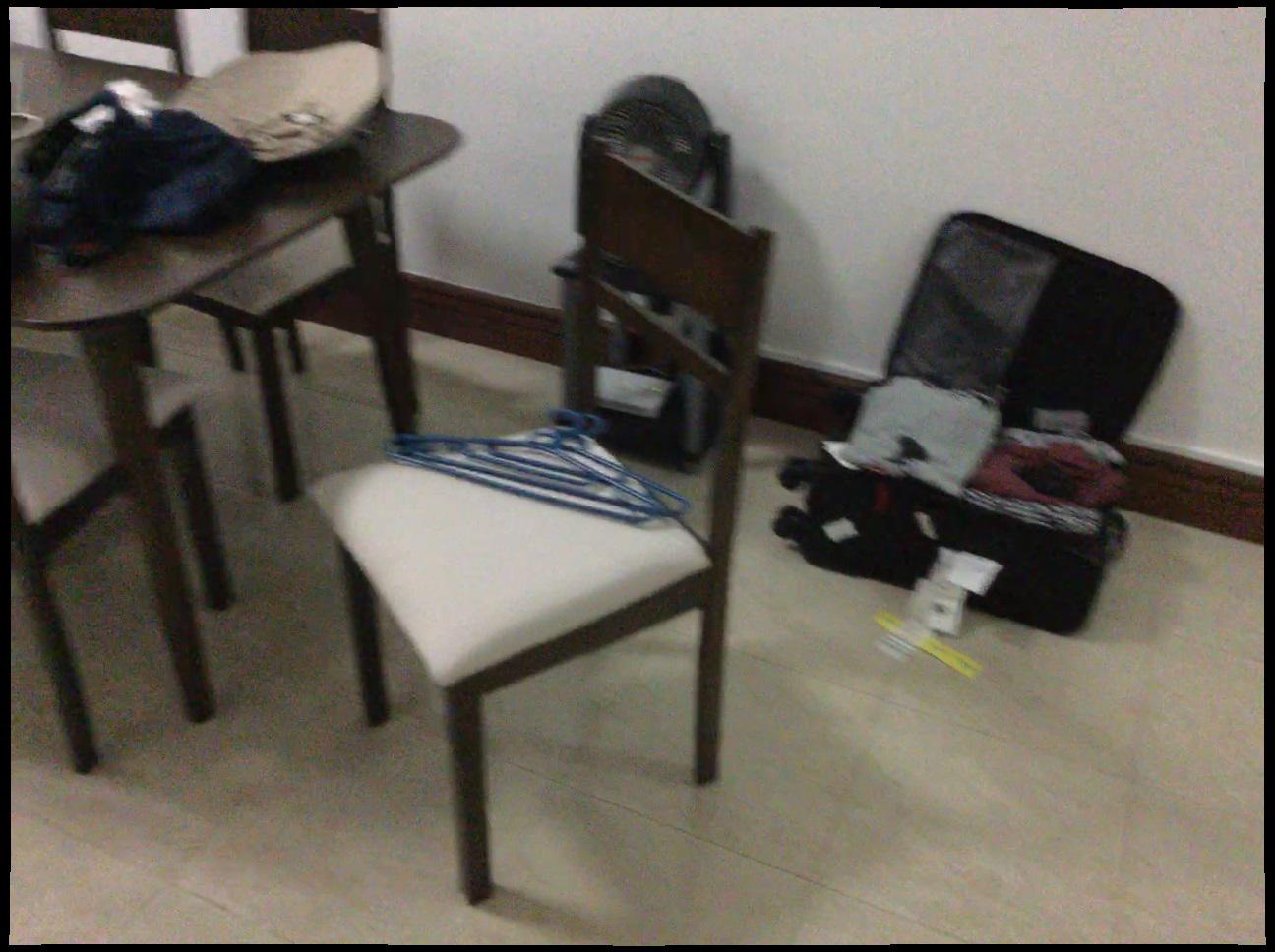} &
        \includegraphics[width=.16\textwidth,height=1.5cm]{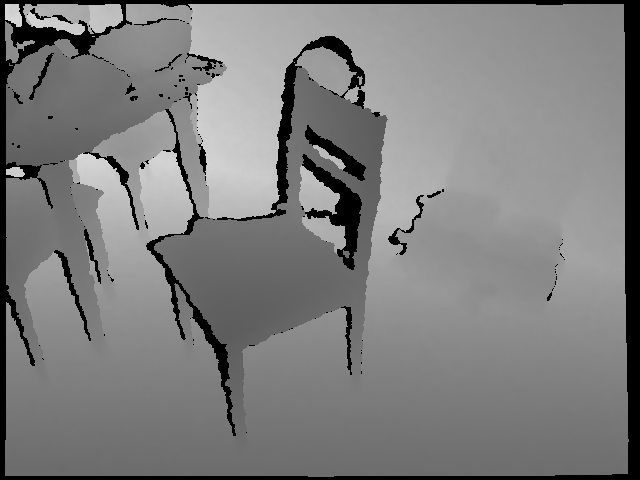} &    \includegraphics[width=.16\textwidth,height=1.5cm]{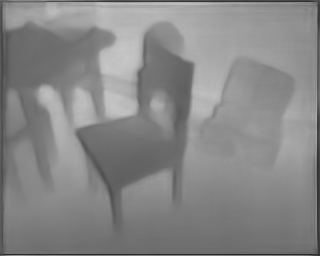} &    \includegraphics[width=.16\textwidth,height=1.5cm]{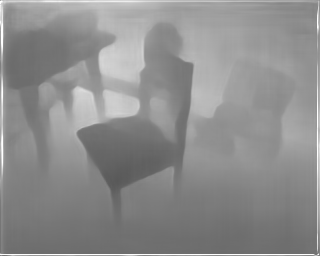} &    \includegraphics[width=.16\textwidth,height=1.5cm]{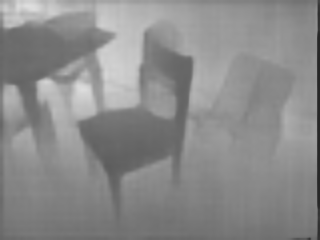} &    \includegraphics[width=.16\textwidth,height=1.5cm]{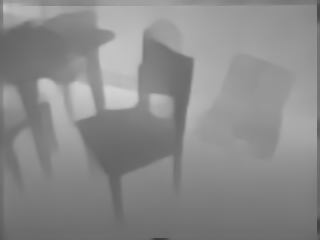} \\         \includegraphics[width=.16\textwidth,height=1.5cm]{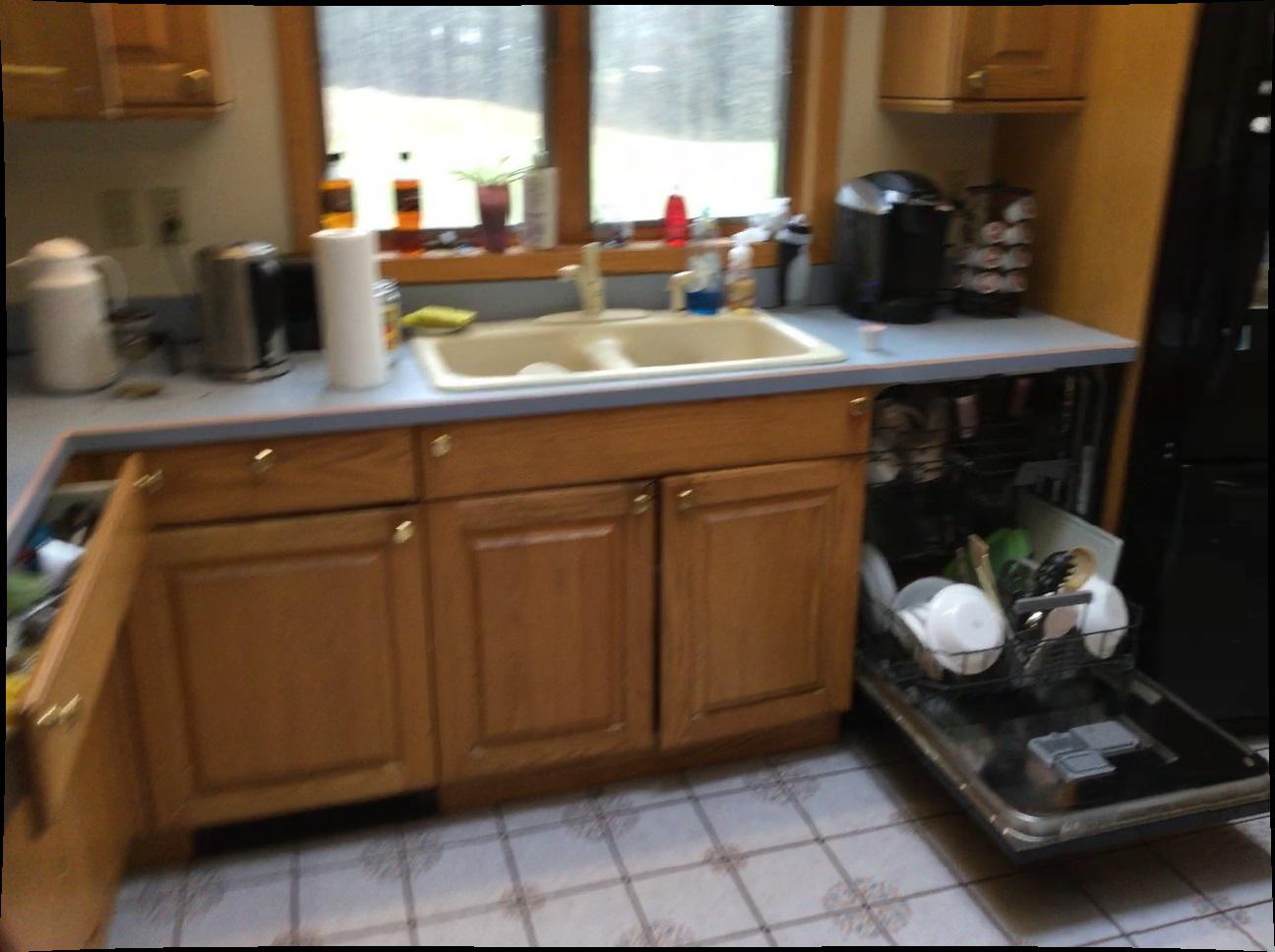} &
        \includegraphics[width=.16\textwidth,height=1.5cm]{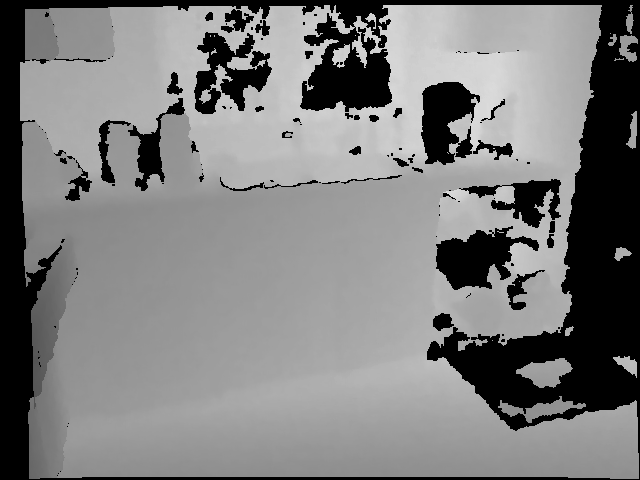} &    \includegraphics[width=.16\textwidth,height=1.5cm]{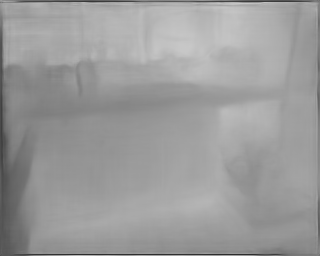} &    \includegraphics[width=.16\textwidth,height=1.5cm]{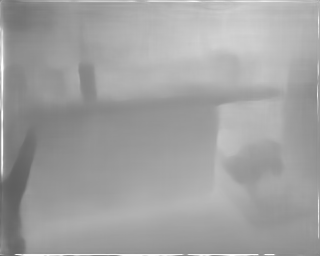} &    \includegraphics[width=.16\textwidth,height=1.5cm]{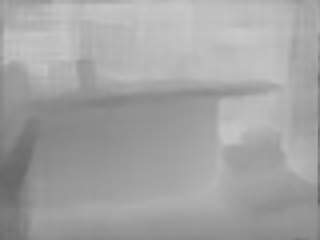} &    \includegraphics[width=.16\textwidth,height=1.5cm]{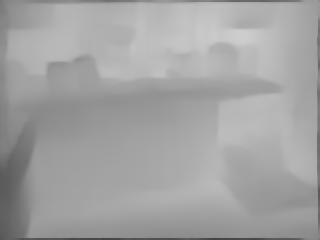} \\         
        Image & GT Depth& MVDepthNet & GPMVSNet & DPSNet & Ours \\
    \end{tabular}
    \caption{Qualitative Performance of our networks on sampled images from ScanNet.}
  \label{fig:device}
 \end{figure}

\vspace{4mm}
\noindent
\textbf{Qualitative results:} In Figure \ref{fig:device} we see that our depth maps are more consistent with respect to ground truth, and respect the geometry of the scene better. For e.g., the lamp in the second row, the chair at the back in the fourth row, the phone in the seventh row and the cabinet in the eight row are qualitatively better than all other methods. Furthermore, we are also able to coherently reconstruct depth where the active depth sensor fails, for e.g. the windows in the second and seventh row and the transparent glass side-table in the sixth row. 



\par\vfill\par

\end{document}